%% file: main.tex
\documentclass[10pt,twocolumn,letterpaper]{article}

\usepackage[pagenumbers]{cvpr} 

\usepackage[accsupp]{axessibility}  
\usepackage{multicol}
\usepackage{multirow}
\usepackage{float}
\usepackage{mathtools}
\usepackage{booktabs}  
\usepackage{arydshln}  
\usepackage{colortbl, xcolor}  
\usepackage{lipsum}
\definecolor{cvprblue}{rgb}{0.21,0.49,0.74}
\usepackage[pagebackref,breaklinks,colorlinks,allcolors=cvprblue]{hyperref}

\title{Interact2Ar: Full-Body Human-Human Interaction Generation via Autoregressive Diffusion Models
\vspace{-3.5mm}
}

\author{
Pablo Ruiz-Ponce$^{
1,2}$,\; Sergio Escalera$^{3}$,\; José García-Rodríguez$^{2}$,\\Jiankang Deng$^{4}$,\; Rolandos Alexandros Potamias$^{1,4}$\\
$^{1}$Huawei Noah’s Ark Lab,\; $^{2}$Universidad de Alicante,\;\\
$^{3}$Universitat de Barcelona and Computer Vision Center,\; $^{4}$Imperial College London\\
\url{https://pabloruizponce.com/papers/Interact2Ar}
}

\begin{document}

\twocolumn[{
\renewcommand\twocolumn[1][]{#1}
\maketitle
\begin{center}
    \centering
    \vspace{-0.6cm}
    \includegraphics[width=0.9\textwidth]{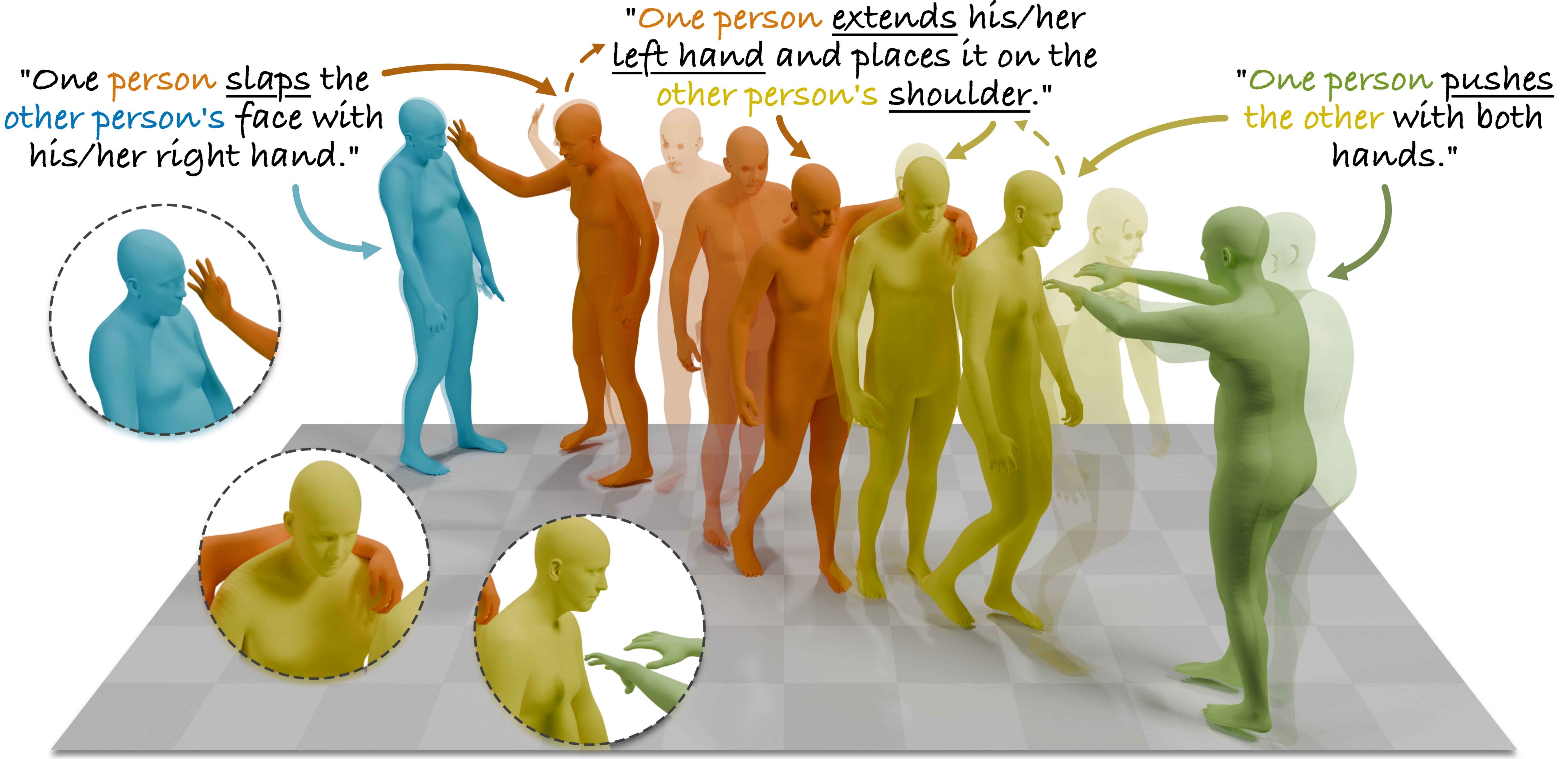}
    \vspace{-0.2cm}
    \captionof{figure}{We introduce \textbf{Interact2Ar}, the first text-conditioned autoregressive diffusion model for generating full-body human-human interactions with detailed hand motions. Our autoregressive model, with a novel memory strategy, enhances the quality of the generated interaction and enables adaptive capabilities, including temporal composition, adaptation to disturbances, and multi-person scenarios.}
    \label{fig:teaser}
\end{center}
}]

\input{sec/00_abstract}    
\input{sec/01_intro}
\input{sec/02_related}
\input{sec/03_method}
\input{sec/04_experimentation}
\input{sec/05_conclusion}

{
    \small
    \bibliographystyle{ieeenat_fullname}
    \bibliography{main}
}

\renewcommand*{\thesection}{\Alph{section}}
\renewcommand*{\thefigure}{\Alph{figure}}
\renewcommand*{\thetable}{\Alph{table}}
\setcounter{section}{0}
\setcounter{table}{0}
\setcounter{figure}{0}
\input{sec/99_suppl}
\end{document}

%% file: sec/00_abstract.tex
\begin{abstract}
Generating realistic human-human interactions is a challenging task that requires not only high-quality individual body and hand motions, but also coherent coordination among all interactants. Due to limitations in available data and increased learning complexity, previous methods tend to ignore hand motions, limiting the realism and expressivity of the interactions. Additionally, current diffusion-based approaches generate entire motion sequences simultaneously, limiting their ability to capture the reactive and adaptive nature of human interactions. To address these limitations, we introduce Interact2Ar, the first end-to-end text-conditioned autoregressive diffusion model for generating full-body, human-human interactions. Interact2Ar incorporates detailed hand kinematics through dedicated parallel branches, enabling high-fidelity full-body generation. Furthermore, we introduce an autoregressive pipeline coupled with a novel memory technique that facilitates adaptation to the inherent variability of human interactions using efficient large context windows. The adaptability of our model enables a series of downstream applications, including temporal motion composition, real-time adaptation to disturbances, and extension beyond dyadic to multi-person scenarios. To validate the generated motions, we introduce a set of robust evaluators and extended metrics designed specifically for assessing full-body interactions. Through quantitative and qualitative experiments, we demonstrate the state-of-the-art performance of Interact2Ar.
\end{abstract}

%% file: sec/01_intro.tex
\vspace{-0.6cm}
\section{Introduction}
\label{sec:intro}
Synthesizing realistic human-human interactions remains a formidable challenge in motion generation~\cite{zhu2023human}. Success requires a model that not only generates high-quality motions for individuals but also coherently accounts for the positioning and dynamics of all participants in the interaction. A primary obstacle limiting the development of such models is the scarcity of high-quality motion capture data~\cite{ruiz2025mixermdm}, a problem more pronounced in this domain than in other generative fields like text and images. Although datasets for single-human motion have driven recent advances \cite{guo2022generating,Plappert2016,AMASS:ICCV:2019}, data for multi-person interaction have remained limited. Previous interaction datasets exhibited significant limitations, lacking interaction diversity~\cite{ng2019you2me,van2011umpm}, sufficient textual annotations~\cite{fieraru2020three,guo2021multi,yin2023hi4d}, and hand motion data~\cite{liang2024intergen}. The recent introduction of the Inter-X dataset~\cite{xu2024inter}, which provides full-body interaction data including detailed hand motions, has been instrumental in addressing these limitations and serves as the basis for training our model, Interact2Ar.

Hand motions provide substantial information in human interactions. Hence, hand modeling is an essential part of non-verbal communication and necessary for capturing contacts between humans. Nevertheless, integrating detailed hand kinematics into human motion generation models presents a significant challenge. The dimensionality of the hands alone exceeds that of the rest of the body. This overhead can introduce more noise than signal, complicating the generation process and leading many models to omit hand information~\cite{tevet2022motionclip, guo2022generating, tevet2023human, tanke2023social, liang2024intergen}. Previous approaches have attempted to incorporate hands by modeling them separately, using parallel~\cite{DSAG} or conditional networks~\cite{ghosh2024remos}. However, these methods suffer from inefficiency or a lack of body context. This problem is compounded in multi-person scenarios where the dimensionality of the interaction scales with the number of individuals. To overcome this, we introduce an end-to-end architecture that leverages cooperative denoisers for effective information flow between individuals. For full-body generation, we employ dedicated branches that generate hands, body, and global trajectory motions in parallel, where each branch is conditioned on a common encoded representation of the motion from the previous denoising step. This unified approach allows for an efficient yet contextually informed generation process, achieving state-of-the-art (SOTA) performance on the Inter-X benchmark.

Despite the architectural advance of specialized heads, current diffusion-based generation pipelines have inherent limitations. These models typically denoise the entire motion sequence at once. While effective for single-human scenarios, this holistic approach struggles to capture the high variability and temporally reactive nature of human interactions, where each person's motions are contingent on the subtle cues of their interaction partner. We therefore introduce the first, to the best of our knowledge, end-to-end text-conditioned autoregressive diffusion model for generating full-body interactions. By generating the motion sequentially, our model can better adapt to the evolving dynamics of the interaction, further improving upon the results obtained by our non-autoregressive approach. To increase the efficiency of memory management, we propose a Mixed Memory approach, where short- and long-term information is included at different frame rates. As a result, a series of additional adaptive interactions are enabled by our model, including temporal motion composition, adaptation to random displacements, and extension to multi-human scenarios beyond dyadic interactions.

To evaluate all these contributions, a robust and detailed evaluation pipeline is needed. In addition to general evaluation metrics, we extend previous quantitative evaluation pipelines to include body-part-specific evaluators that provide detailed insights about motion quality. We also increased their robustness by retraining all evaluators with the global positions of the interactants instead of rotation-based representations. Finally, we included jerk-based metrics to evaluate the smoothness of transitions, which is particularly important for assessing the new downstream capabilities of our autoregressive model.

\noindent The main contributions of this paper are as follows:
\begin{itemize}
    \item We propose an end-to-end diffusion model for generating human-human interactions with detailed hand motions, achieving state-of-the-art performance on Inter-X.
    \item We introduce an autoregressive diffusion pipeline for human-human interaction generation that improves interaction quality through sequential motion synthesis with efficient memory management. Its adaptability enables downstream tasks, including temporal composition, disturbance adaptation, and multi-human generation.
    \item We extend quantitative evaluation by proposing more robust and specialized evaluators, enabling more reliable and informative assessment.
\end{itemize}

%% file: sec/02_related.tex
\section{Related Work}
\label{sec:related}
\textbf{Human Motion Generation.} The field of human motion generation has experienced remarkable growth in recent years~\cite{zhu2023human}, driven by the introduction of large-scale datasets~\cite{AMASS:ICCV:2019,guo2022generating,h36m_pami,cai2022humman,xu2023towards,shahroudy2016ntu,guo2020action2motion,BABEL:CVPR:2021} and novel generative architectures~\cite{guo2020action2motion, tevet2022motionclip, tevet2023human, guo2024momask}. However, such rapid expansion has introduced several challenges that distinguish this area from more mature research domains. One issue is the lack of standardized data representations~\cite{SMPL:2015, SMPL-X:2019, meng2025absolute, guo2022generating, liang2024intergen, yu2025socialgen}. While certain representations are designed to facilitate the computation of kinematic losses and to improve model learning~\cite{guo2022generating, liang2024intergen, yu2025socialgen}, they suffer from memory inefficiency. Another challenge is the integration of hand kinematics, which are frequently omitted due to their high dimensionality and susceptibility to noise~\cite{tevet2022motionclip, guo2022generating, tevet2023human}, despite their critical role in expressive motion synthesis. Additionally, there is no established generative paradigm that consistently outperforms others. Diffusion-based approaches~\cite{tevet2023human, kim2023flame, zhang2024motiondiffuse, chen2023executing, yuan2023physdiff, zhang2023remodiffuse, azadi2023make, zhou2024emdm, huang2024stablemofusion} typically deliver superior motion quality and greater adaptability to diverse conditions, but at the cost of slower inference and more intensive training requirements. Conversely, quantized representation methods~\cite{guo2024momask, yuan2024mogents, pinyoanuntapong2024mmm, guo2022tm2t, zhang2023generating, zhong2023attt2m, jiang2023motiongpt, zhang2024motiongpt} offer faster training and inference, and their learned discrete tokens enable robust motion quality with minimal tuning, though with reduced flexibility in conditioning and control. Finally, evaluation remains a significant challenge. While standard metrics provide a general assessment of model performance, recent studies~\cite{meng2025rethinking, meng2025absolute} have exposed their limitations. These studies reveal biases toward certain architectures and question their reliability as indicators of true motion quality.

\noindent\textbf{Human-Human Interaction Generation.} Generating human-human interactions substantially amplifies the challenges inherent in single-person motion synthesis~\cite{sui2025survey,fan20253d}. While progress has been facilitated by different datasets~\cite{yin2023hi4d,guo2021multi,fieraru2020three,ng2019you2me,van2011umpm}, with the most recent ones being InterHuman~\cite{liang2024intergen} and Inter-X~\cite{xu2024inter}, the modeling complexity is considerably higher than for single-human motions. Unlike single-person scenarios, models must learn intricate spatio-temporal dependencies to ensure that each agent's motion remains coherent with respect to others. Capturing the full distribution of human interactions represents a formidable challenge that current datasets only partially address. Inter-X has enabled new research directions in full-body interaction generation, including detailed hand articulation, which is frequently omitted due to its modeling complexity~\cite{tanke2023social, liang2024intergen, ruiz2024in2in}. Several approaches have tackled interaction modeling through different architectural paradigms~\cite{tanke2023social, shafir2023human, cai2024digital, yu2025socialgen, tanke2025dyadic, xu2025multi, wang2024intercontrol}. InterGen~\cite{liang2024intergen} and in2IN~\cite{ruiz2024in2in} employed diffusion models with cooperative denoising mechanisms to capture inter-agent dependencies. More recently, InterMask~\cite{javed2024intermask} introduced residual VQ-VAE masked transformers for interaction generation, achieving state-of-the-art performance on the Inter-X benchmark. Our work proposes a novel diffusion-based model with body-specific heads capable of synthesizing high-fidelity human-human interactions with detailed hand dynamics, achieving state-of-the-art results on the Inter-X benchmark.

\noindent\textbf{Autoregressive Diffusion Models for Human Motion Generation.} A key limitation of diffusion models in human motion generation is their tendency to generate entire motion sequences in a single forward pass~\cite{tevet2023human}. While this approach may suffice for simple, isolated single-person scenarios, it fails to capture the dynamic, responsive nature of humans interacting in the real world. Recently, autoregressive diffusion models have been introduced to generate short motion segments conditioned on a temporal context window~\cite{camdm,shi2024interactive, zhao2024dartcontrol, xiao2025motionstreamer}. This paradigm has enabled single-person motion generation systems to exhibit greater responsiveness to diverse inputs, such as environmental context~\cite{tevet2025closd} and external physical forces~\cite{zhang2025primalphysicallyreactiveinteractive}. More recently, autoregressive diffusion has shown promising results in reaction generation tasks~\cite{ji2025towards, cen2025ready_to_react}. To the best of our knowledge, our work presents the first end-to-end text-conditioned autoregressive diffusion model for full-body human-human interactions. The rationale for adopting this paradigm is that autoregressive generation enables dynamic adaptation of each individual's motion based on their partner's movements and the evolving interaction context. This approach yields models capable of capturing the inherent variability and responsiveness of human-human interactions. We demonstrate this capability through a series of downstream tasks that evaluate the model's adaptability to dynamic interaction scenarios.

%% file: sec/03_method.tex
\section{Method}
\label{sec:method}
Interact2Ar generates full-body human-human interactions $x$ conditioned on textual descriptions $c$. We present a diffusion model (\cref{sec:method:architecture}) using cooperative denoisers to facilitate information flow between interactants, along with specialized branches for generating the different body parts. We further improve our denoiser architecture with an autoregressive pipeline (\cref{sec:method:ar}) that replaces full motion generation with a step-wise approach where the denoiser predicts sub-motions while retaining a prefix of predefined length from the previous sub-motions as memory. To improve memory management, we propose a mixed strategy (\cref{sec:method:memory}) that efficiently provides both short-term and long-term information. All these methodological contributions provide adaptive capabilities to Interact2Ar that enable additional downstream applications (\cref{sec:method:applications}).

\subsection{Dataset \& Motion Representation}
\label{sec:method:data}
Inter-X~\cite{xu2024inter} is the most suitable dataset for training Interact2Ar as it contains 11K full-body interactions performing 40 different actions with detailed textual descriptions.

Redundant representations, where certain features can be derived from others (e.g., joint velocities from positions), have become common in human motion modeling, as they facilitate learning and accelerate the computation of certain losses during training. However, this approach increases feature dimensionality, which becomes especially problematic when representing two individuals with detailed hand articulations. Furthermore, redundant representations are suboptimal for diffusion model evaluation pipelines, as they introduce negative biases~\cite{meng2025rethinking}. To address these issues, we rely exclusively on SMPL-X parameters~\cite{SMPL-X:2019} to represent interactions. A dyadic interaction $x = \{\prescript{a}{}{x}, \prescript{b}{}{x}\}$ between individuals $a$ and $b$ is composed of individual poses, where each $\prescript{i}{}{x}$ is represented as $(r, \varphi, \theta_{\text{body}}, \theta_{\text{hands}})$, with $r\in\mathbb{R}^3$ denoting the root translation, and $\varphi \in \mathbb{R}^6$, $\theta_{\text{body}}\in\mathbb{R}^{21 \times 6}$, $\theta_{\text{hands}}\in\mathbb{R}^{30 \times 6}$ representing the root and joint rotations using the continuous 6D representation~\cite{zhou2019continuity}. Since the dataset exhibits limited body shape diversity, we normalize all shapes to a neutral body configuration.

\begin{figure*}[!htp]
  \centering
    \includegraphics[width=\textwidth]{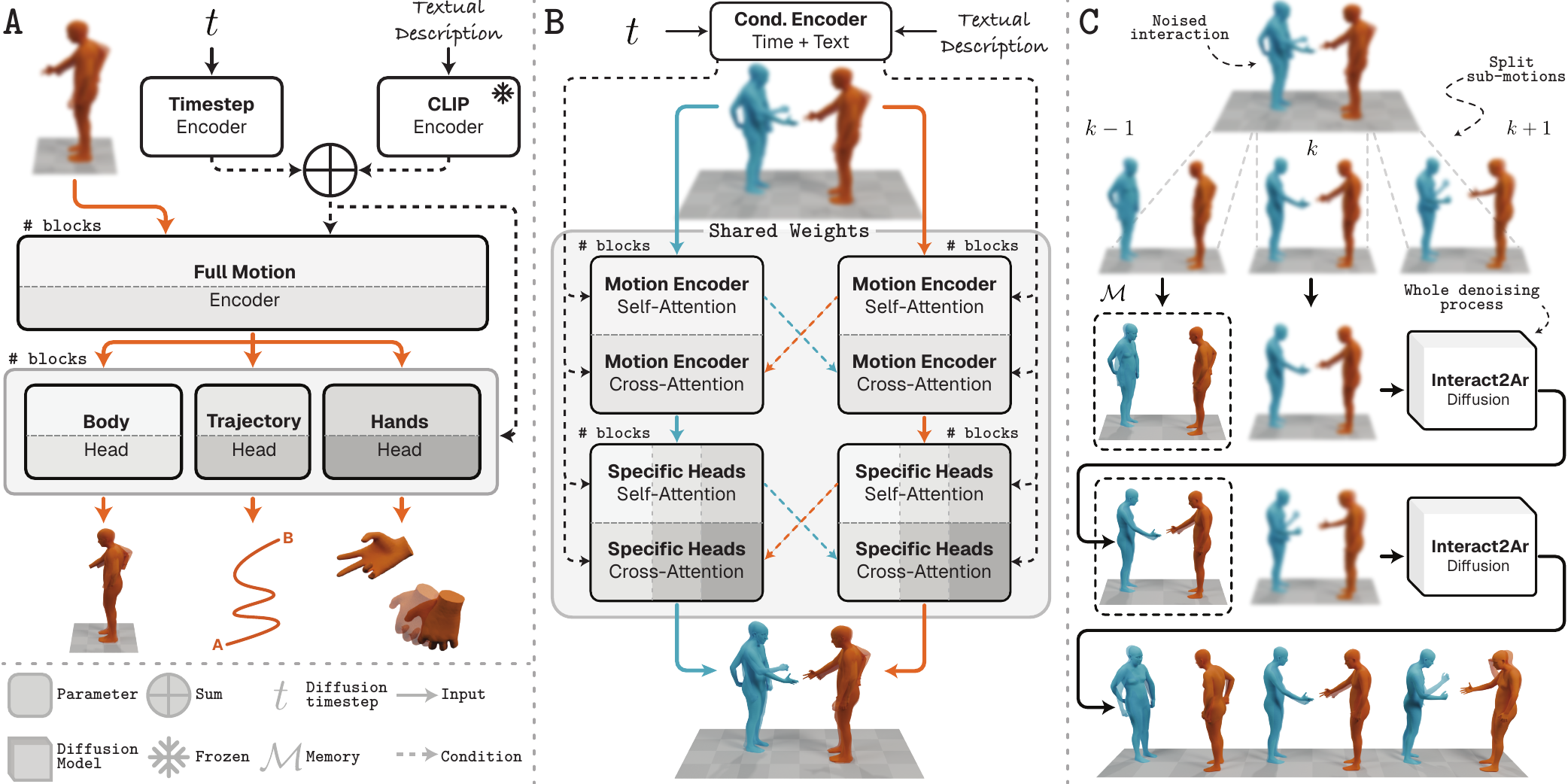}
    \vspace{-0.4cm}
  \caption{\textbf{A) Multi-Head Denoiser}: An encoder feeds noised motion and condition information to specialized heads for body, trajectory, and hands denoising. \textbf{B) Cooperative Denoisers}: Parallel streams with shared weights generate each interactant while cross-attention shares inter-personal information. \textbf{C) Autoregressive}: Sequential generation of sub-motions conditioned on previously generated frames.}
  \label{fig:architecture}
    \vspace{-0.4cm}
\end{figure*}

\subsection{Full-Body Interaction Generation}
\label{sec:method:architecture}

For generating human-human interactions, we employ diffusion models for their quality and adaptability~\cite{xiao2022DDGAN, dhariwal2021diffusion, ruiz2025mixermdm}. In this paradigm, we incrementally add Gaussian noise $\mathcal{N}(0,1)$ to the motions and train a denoising network that learns to incrementally remove the added noise. The denoiser takes as input the textual condition $c$ and the partially noised interaction $x^{t}$, and predicts $\hat{x}^{t-1}$, where $t$ represents the step in the diffusion chain and the $\hat{\text{hat}}$ notation denotes a prediction made by our model. We follow~\cite{ramesh2022hierarchical, tevet2023human} and directly predict $\hat{x}^{0}$ at each denoising step, instead of $\hat{x}^{t-1}$, to enable the calculation of kinematic losses.

\noindent\textbf{Architecture.} We build upon the cooperative denoiser architecture proposed by InterGen \cite{liang2024intergen}. This transformer-encoder architecture processes each individual in parallel through separate streams that share weights. To flow information between individuals, hidden states are exchanged between streams via cross-attention. This design choice enables a more compact network with fewer trainable parameters while preserving interaction information.

However, this architecture predicts entire pose sequences without body-part specialization, which can hinder accurate modeling of high-dimensional representations, particularly for complex hand articulations. We address this by introducing an initial encoding module that maps noised motion into a latent representation, which feeds three specialized denoising heads for global trajectory, body poses, and hand poses. This design enables body-specific modules that leverage full motion information for coherent predictions while allowing parallel computation across heads. The architecture is illustrated in \cref{fig:architecture}.

\noindent\textbf{Training Losses.} We adopt a combination of commonly used losses for training diffusion models in the task of human-human interaction generation~\cite{liang2024intergen,ruiz2024in2in}:
\vspace{-1mm}
\begin{equation}
\begin{aligned}
    \mathcal{L}_{\text{total}} &= \lambda_{\text{repr}} \mathcal{L}_{\text{repr}}(x,\hat{x}) + \lambda_{\text{orient}} \mathcal{L}_{\text{orient}}(r,\hat{r}) \\[0.3em]
    &\quad + \lambda_{\text{pos}} \mathcal{L}_{\text{pos}}(p, \hat{p}) + \lambda_{\text{vel}} \mathcal{L}_{\text{vel}}(v, \hat{v}) \\[0.3em]
    &\quad + \lambda_{\text{foot}} \mathcal{L}_{\text{foot}}(f, \hat{f}) + \lambda_{\text{dist}} \mathcal{L}_{\text{dist}}(d, \hat{d}),
\end{aligned}
\end{equation}
\vspace{-1mm}

\noindent where $\mathcal{L}_{\text{repr}}$ is the $\ell_2$ loss between raw SMPL-X representations, $\mathcal{L}_{\text{orient}}$ penalizes root orientation errors, and the remaining four terms are kinematic losses computed via forward kinematics (FK). To compute these, we pass predicted SMPL-X parameters through a differentiable FK layer to obtain joint positions $p = \text{FK}(x)$ and $\hat{p} = \text{FK}(\hat{x})$, from which we compute: $\mathcal{L}_{\text{pos}}$ for global joint positions, $\mathcal{L}_{\text{vel}}$ for joint velocities, $\mathcal{L}_{\text{foot}}$ for foot contacts, and $\mathcal{L}_{\text{dist}}$ for pairwise joint distance maps between individuals. Weighting coefficients $\{\lambda_i\}$ were tuned through grid search. Additional details are provided in the Supplementary Material.

\subsection{Autoregressive Interaction Generation}
\label{sec:method:ar}
Unlike single-human motion generation, human-human interaction requires a high level of adaptability to respond to changes in the interaction state. Traditional diffusion methods employed for motion generation cannot achieve this level of adaptability as they predict the entire sequence at once. We extend our denoiser by proposing an autoregressive diffusion pipeline for human-human interaction following the advancements in single-human scenarios~\cite{tevet2025closd}.

An interaction $x$ of total length $N$ is decomposed into a series of contiguous non-overlapping sub-motions:
\vspace{-1mm}
\begin{equation}
    x = \bigcup_{k=0}^{K-1} {x_{kn:(k+1)n}},
\end{equation}
\vspace{-3mm}

\noindent where $K = \lceil N/n \rceil$ denotes the number of sub-motions, and $n$ represents the generation window, i.e., the number of frames the denoiser can output in a single forward pass. At each generation step $k$, the denoiser predicts the next sub-motion conditioned on a short-term memory buffer $\mathcal{M}^s$ of the $m_s$ most recent previously generated frames:
\vspace{-1mm}
\begin{equation}
    \mathcal{M}_k^s = {x^0_{kn-m_s:kn}}
\end{equation}
\vspace{-1mm}
The denoiser then predicts:
\vspace{-1mm}
\begin{equation}
    \hat{x}^{0}_{kn:(k+1)n} = G(x^t_{kn:(k+1)n}, \mathcal{M}_k^s, c, t),
\end{equation}
where $x^t_{kn:(k+1)n}$ is the noised sub-motion at diffusion timestep $t$, $c$ is the textual condition, and $t$ is the current diffusion timestep. The memory size $m_s$ controls the temporal context available to the model, allowing it to maintain coherence with past motions while generating new frames.

\begin{figure}[!htbp]
  \centering
    \includegraphics[width=\columnwidth]{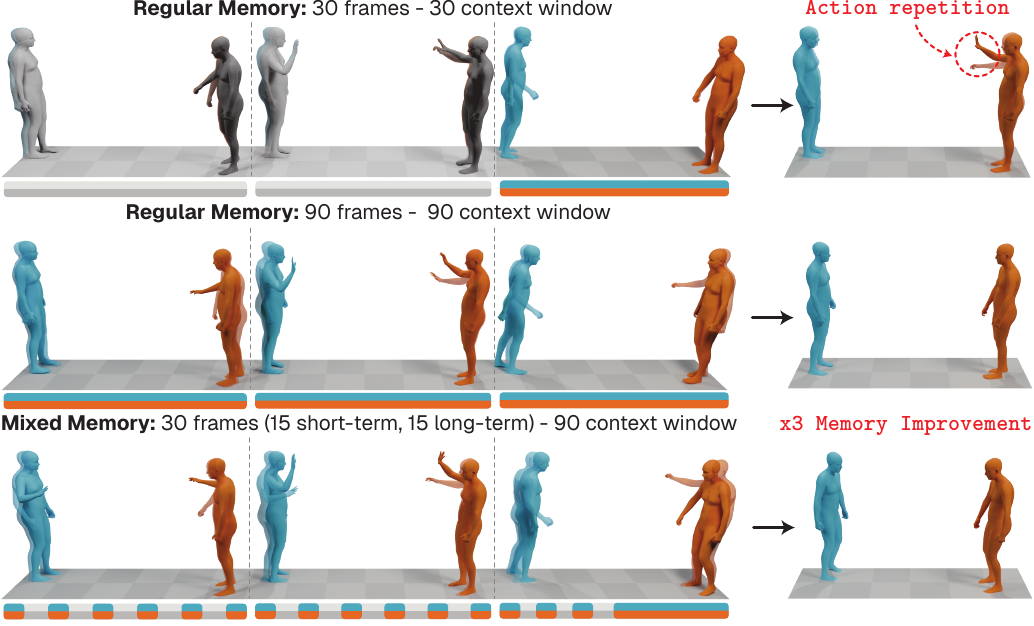}
    \vspace{-0.6cm}
    \caption{\textbf{Mixed Memory} enables access to both detailed short-term information, facilitating seamless transitions, along with long-term context, avoiding action repetition in long interactions. Our proposed Mixed Memory overcomes the limitations of regular context memory, providing up to a $\times$3 reduction in memory size.}
  \label{fig:mixed}
    \vspace{-0.4cm}
\end{figure}

\noindent\textbf{Mixed Memory.}
\label{sec:method:memory}
A key limitation of this approach is the constrained short-term memory capacity. Recent works typically employ prediction windows of 20 to 40 frames, with memory windows of a similar range~\cite{tevet2025closd, zhang2025primalphysicallyreactiveinteractive}. While this can be sufficient for short or repetitive motions, our observations indicate that for long-duration interactions (\cref{fig:mixed}), such limited memory adversely affects output quality, resulting in repetitive motion artifacts due to insufficient historical context.

To address this limitation, we augment the short-term memory with a long-term component $\mathcal{M}^l$. We maintain a temporally downsampled history spanning a substantially longer window by retaining frames at intervals of $\delta$ over a temporal window of $m_l$ frames, where $m_l \gg m_s$:
\vspace{-1mm}
\begin{equation}
    \mathcal{M}_k^l = \{x^0_{kn-m_l+i\delta} \mid i = 0, 1, \ldots, \lfloor m_l/\delta \rfloor\}
\end{equation}
The complete memory buffer is constructed by concatenating both components:
\vspace{-1mm}
\begin{equation}
    \mathcal{M}_k = \{\mathcal{M}_k^l, \mathcal{M}_k^s\}
\end{equation}
\vspace{-1mm}
The denoiser now predicts conditioned on Mixed Memory:
\vspace{-1mm}
\begin{equation}
    \hat{x}^{0}_{kn:(k+1)n} = G(x^t_{kn:(k+1)n}, \mathcal{M}_k, c, t)
\end{equation}
Our dual-memory architecture enables the model to leverage full-framerate immediate context for seamless transitions and long-range temporal information at a downsampled framerate more efficiently than covering the whole context window with simple memory. The downsampling rate $\delta$ and long-term window size $m_l$ are hyperparameters that control the trade-off between memory coverage and computational efficiency. A graphical illustration of this process is shown in \cref{fig:mixed}.

\subsection{Adaptive Interactions}
\label{sec:method:applications}
Thanks to our proposed autoregressive pipeline with a Mixed Memory strategy, Interact2Ar presents a series of adaptability capabilities that enable several downstream interaction applications. A quantitative and qualitative analysis of these capabilities can be found in \cref{sec:experiments:additional}, and a comparison with previous methods in the Supplementary Video.

\noindent\textbf{Temporal Motion Composition.} The autoregressive generation naturally enables temporal composition of different actions with seamless transitions when switching between textual prompts. Because the denoiser conditions on the memory buffer $\mathcal{M}$ from previously generated sub-motions, it natively handles compositional transitions during generation without requiring offline post-processing or suffering from the global positioning misalignments common in inpainting-based approaches.

\noindent\textbf{Real-Time Disturbance Adaptation.} The step-wise generation paradigm enables real-time adaptation to external disturbances. By generating sub-motions of length $n$ conditioned on recent history, the model can adapt to significant state perturbations between generation steps (e.g., sudden position changes, unexpected contact events), demonstrating reactive generation beyond pre-planned sequences.

\noindent\textbf{Sequential Multi-Person Interactions.} Combining temporal composition with disturbance adaptation enables scenarios where one individual sequentially interacts with multiple partners. After completing an interaction with one partner, a new interactant can be introduced with a different prompt, and the memory-conditioned generation ensures smooth transitions between sequential dyadic interactions without requiring simultaneous multi-person modeling.

%% file: sec/04_experimentation.tex
\section{Experimentation}
\label{sec:experimentation}
\textbf{Implementation details.} The cooperative denoiser uses 8 transformer blocks with 8 attention heads (latent dimension 512, feed-forward dimension 1024) for the motion encoder and both body and hand pose heads, while the trajectory head uses 4 blocks with 4 heads (latent dimension 256, feed-forward dimension 512). For diffusion steps, the full model uses 1000 steps with DDIM-50 sampling, while the autoregressive version achieves the best empirical results with only 10 steps. All models were trained for 5000 epochs using EMA and AdamW (learning rate $1 \times 10^{-4}$, weight decay $2 \times 10^{-5}$, batch size 128). Additional implementation details are provided in the Supplementary materials.

\noindent\textbf{Evaluation Metrics.} We adopt the widely used metrics \cite{guo2022generating} for measuring generation quality and textual alignment: R-Precision (R-Prec.), Frechet Inception Distance (FID), MultiModal Distance (MM Dist), Diversity, and MultiModality (MModality). To quantitatively evaluate the adaptive nature of our model, we employ the Peak Jerk (PJ) and Area Under the Jerk (AUJ) metrics from \cite{barquero2024seamless} to assess the smoothness transitioning from different states. We generate 64 sequences of 8 temporally concatenated motions each and calculate the PJ and AUJ across the complete sequences, implementing concatenation through inpainting for models lacking native support. Following Inter-X, all metrics are calculated per interaction rather than per individual to faithfully evaluate the quality of the interactions.

\noindent\textbf{Full Body Evaluation.} 
As we experimentally show in \cref{tab:robustness}, the original evaluator from Inter-X has serious limitations in recognizing degradation in the interactions. It maintains the same performance even when serious degradation is applied to the trajectory or when the trajectories of the individuals are swapped. To address this, we retrained the original evaluator using only joint positions instead of rotations to minimize bias and improve evaluation quality~\cite{meng2025rethinking}. Given the importance of the global position of an individual with respect to another, we opt to use global joint coordinates instead of relative. To enhance the interpretability of full-body evaluation, we train three distinct evaluators for different body components: one for all joints (as in the original benchmark), one for body joints only, and one for hand joints only. In \cref{tab:robustness}, we can observe that our new evaluators recognize and penalize much more serious degradations than previous evaluators.

\subsection{Comparison to State-of-the-art Approaches}

\noindent\textbf{Quantitative Evaluation}. \cref{tab:results} presents the quantitative comparison of Interact2Ar against previous state-of-the-art (SOTA) methods across different aspects and levels of detail. In terms of motion quality, both the non-autoregressive and autoregressive versions of Interact2Ar surpass the previous SOTA, InterMask. Examining the results more closely reveals improvements in the evaluation metrics for body and hands independently. These results demonstrate that our proposed cooperative denoiser architecture handles this full-body motion representation substantially better than previous approaches. Furthermore, the autoregressive version consistently outperforms the non-autoregressive version across all metrics, highlighting the advantages of this paradigm for modeling interactions. Please note that MultiModality metric tends to yield lower values for models that perform better and align more closely to the textual description provided~\cite{ruiz2024in2in}.

\begin{figure*}[!htb]
  \centering        
  \includegraphics[width=0.9\textwidth]{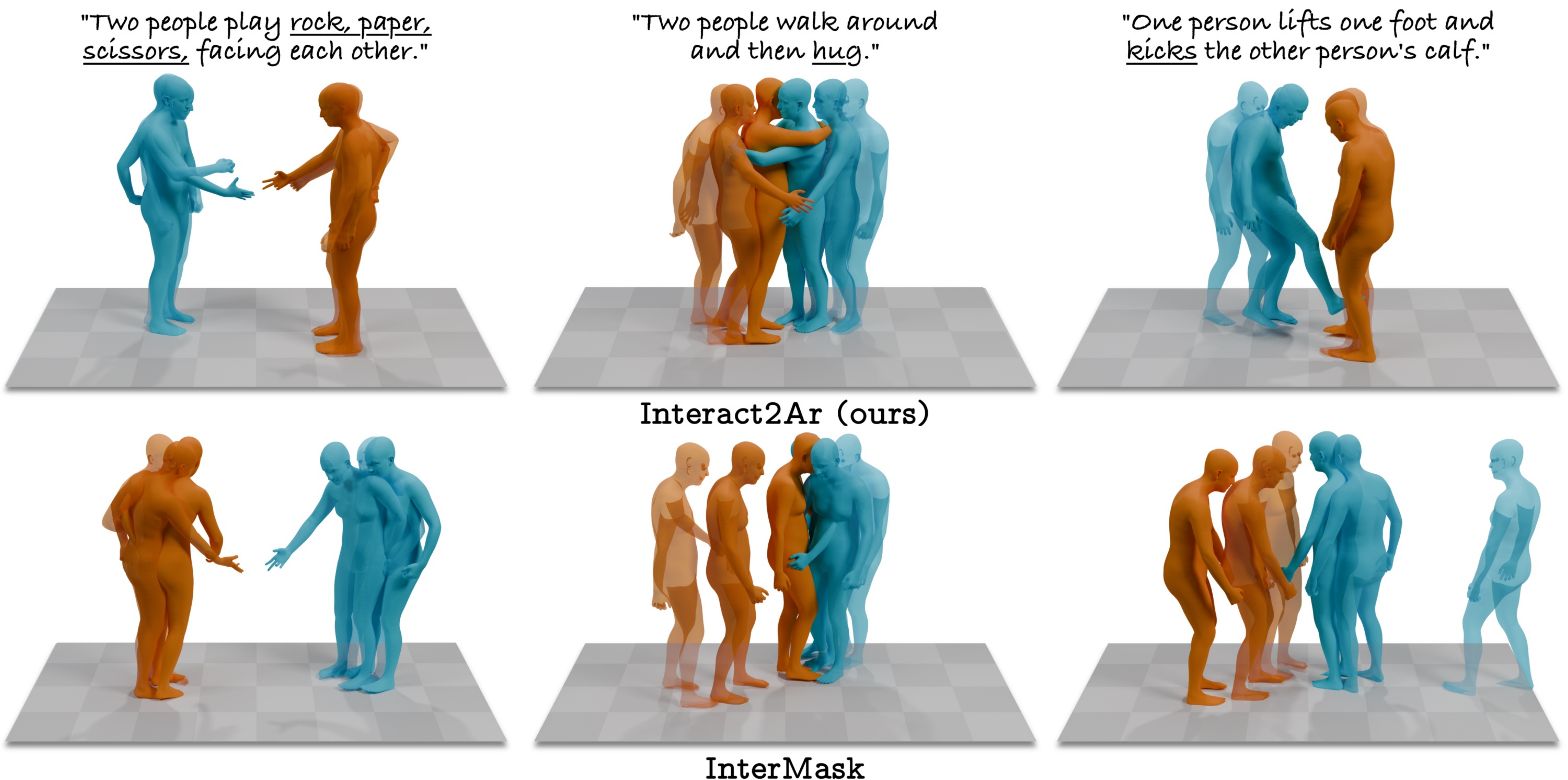}
  \vspace{-0.2cm}
  \caption{\textbf{Interact2Ar comparison with SOTA}. Our method Interact2Ar (top) generates higher-quality interactions with improved alignment to textual descriptions and more realistic hand motions in comparison to the previous SOTA InterMask (bottom).}
  \label{fig:qualitative:hhi}
  \vspace{-0.4cm}
\end{figure*}

\begin{table}[!htbp]
\centering
\small
\setlength{\tabcolsep}{2pt}
\begin{tabular}{c|cc|cc}
\toprule
\multirow{2}{*}{Method} & \multicolumn{2}{c|}{Prev. Evaluator~\cite{xu2024inter}} & \multicolumn{2}{c}{Ours} \\
& R-Prec. $\uparrow$ & FID $\downarrow$ & R-Prec. $\uparrow$ & FID $\downarrow$ \\
\midrule
\cellcolor{blue!5}\textit{Ground Truth} & \cellcolor{blue!5}$0.739^{\pm.00}$ & \cellcolor{blue!5}$0.001^{\pm.00}$ & \cellcolor{blue!5}$0.740^{\pm.00}$ & \cellcolor{blue!5}$0.002^{\pm.00}$ \\
\midrule
Interact2Ar & $0.737^{\pm.00}$ & $0.148^{\pm.01}$ & $0.773^{\pm.00}$ & $0.277^{\pm.01}$ \\
+10\% noise & $0.307^{\pm.00}$ & $38.35^{\pm.12}$ & $0.211^{\pm.00}$ & $74.58^{\pm.14}$ \\
+10\% noise traj. & $0.737^{\pm.00}$ & $0.122^{\pm.00}$ & $0.249^{\pm.00}$ & $62.05^{\pm.12}$ \\
+ traj. swap & $0.729^{\pm.00}$ & $0.165^{\pm.01}$ & $0.558^{\pm.00}$ & $8.65^{\pm.05}$ \\
\bottomrule
\end{tabular}
\vspace{-0.2cm}
\caption{\textbf{Evaluator robustness comparison.} Our evaluator demonstrates superior sensitivity to motion quality degradations compared to the previous one. Tested degradations: noise on full representation, noise on trajectory only, and trajectory swapping.}
\label{tab:robustness}
\end{table}
\vspace{-0.2cm}
\noindent\textbf{Qualitative Evaluation.} Beyond quantitative evaluation, qualitative comparisons better illustrate the differences in quality between previous methods and ours. Our method clearly generates higher-quality interactions with improved alignment to textual descriptions and more realistic hand motions (\cref{fig:qualitative:hhi}). The Supplementary Video provides additional qualitative comparisons, allowing for a more comprehensive appreciation of the temporal and spatial coherence and overall realism achieved by our method.

\begin{figure}[!htbp]
  \centering
    \includegraphics[width=\linewidth]{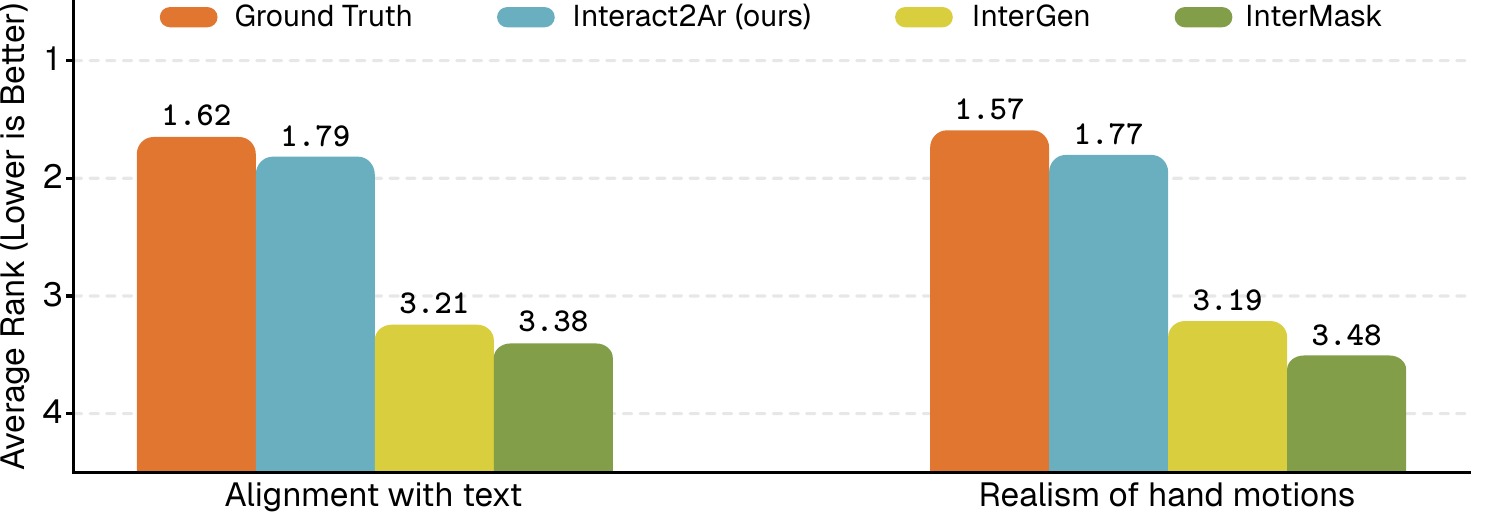}
    \vspace{-0.6cm}
    \caption{\textbf{User study} with the average ranking of 35 participants evaluating the text alignment and hand quality of 10 interactions.}  \label{fig:userstudy}
    \vspace{-0.5cm}
\end{figure}

\noindent\textbf{User Study.} To assess generation quality, we conducted a user study with 35 participants. Users ranked 10 videos, each containing one interaction extracted from: ground truth, our model, InterMask, and InterGen. For each video, participants ranked the generations based on (1) overall quality and alignment with the textual description, and (2) realism of hand motions. As shown in \cref{fig:userstudy}, our method clearly outperforms previous approaches across both criteria and approaches ground-truth quality.    
\vspace{-0.1cm}

\begin{table*}[!htbp]
\centering
\small
\setlength{\tabcolsep}{3.5pt}
\begin{tabular}{c|c|ccccccc|cc}
\toprule
& \multirow{2}{*}{Methods} & \multicolumn{3}{c}{R-Precision $\uparrow$} & \multirow{2}{*}{FID $\downarrow$} & \multirow{2}{*}{MM Dist $\downarrow$} & \multirow{2}{*}{Diversity $\rightarrow$} & \multirow{2}{*}{MModality $\uparrow$} & \multirow{2}{*}{PJ $\uparrow$} & \multirow{2}{*}{AUJ $\downarrow$} \\
& & Top 1 & Top 2 & Top 3 & & & & & & \\
\midrule
\multirow{6}{*}{\rotatebox{90}{Full}} 
& \cellcolor{blue!5}\textit{Ground Truth} & \cellcolor{blue!5}$0.431^{\pm.00}$ & \cellcolor{blue!5}$0.631^{\pm.00}$ & \cellcolor{blue!5}$0.740^{\pm.00}$ & \cellcolor{blue!5}$0.002^{\pm.00}$ & \cellcolor{blue!5}$3.318^{\pm.01}$ & \cellcolor{blue!5}$8.973^{\pm.10}$ & \cellcolor{blue!5}-- & \cellcolor{blue!5}$0.021^{\pm.00}$ & \cellcolor{blue!5}$3.944^{\pm.00}$ \\
\cdashline{2-11}
& T2M~\cite{guo2022generating} & $0.211^{\pm.00}$ & $0.339^{\pm.00}$ & $0.434^{\pm.00}$ & $9.079^{\pm.11}$ & $5.766^{\pm.02}$ & $7.994^{\pm.08}$ & $0.789^{\pm.03}$ & $1.889^{\pm.12}$ & $124.9^{\pm1.9}$ \\
& InterGen~\cite{liang2024intergen} & $0.411^{\pm.01}$ & $0.608^{\pm.00}$ & $0.721^{\pm.00}$ & $0.874^{\pm.02}$ & $3.618^{\pm.01}$ & $\underline{8.876}^{\pm.07}$ & $\textbf{3.345}^{\pm.07}$ & $2.132^{\pm.12}$ & $84.16^{\pm3.1}$ \\
& InterMask~\cite{javed2024intermask} & $0.415^{\pm.00}$ & $0.607^{\pm.00}$ & $0.722^{\pm.00}$ & $0.671^{\pm.02}$ & $3.487^{\pm.01}$ & $8.654^{\pm.06}$ & $1.686^{\pm.06}$ & $2.328^{\pm.21}$ & $61.74^{\pm.50}$ \\
\cdashline{2-11}
& \cellcolor{blue!5}Interact2Ar* & $\underline{0.436}^{\pm.00}$ & $\underline{0.643}^{\pm.00}$ & $\underline{0.757}^{\pm.00}$ & $\underline{0.556}^{\pm.02}$ & $\underline{3.246}^{\pm.01}$ & $\textbf{8.916}^{\pm.07}$ & $\underline{2.743}^{\pm.05}$ & $2.110^{\pm.21}$ & $54.97^{\pm.67}$ \\
& \cellcolor{blue!10}Interact2Ar & $\textbf{0.453}^{\pm.00}$ & $\textbf{0.661}^{\pm.00}$ & $\textbf{0.773}^{\pm.00}$ & $\textbf{0.277}^{\pm.01}$ & $\textbf{3.095}^{\pm.01}$ & $9.305^{\pm.07}$ & $1.427^{\pm.04}$ & $\textbf{0.136}^{\pm.00}$ & $\textbf{8.837}^{\pm.17}$ \\
\midrule
\multirow{6}{*}{\rotatebox{90}{Body}} 
& \cellcolor{blue!5}\textit{Ground Truth} & \cellcolor{blue!5}$0.462^{\pm.01}$ & \cellcolor{blue!5}$0.655^{\pm.00}$ & \cellcolor{blue!5}$0.758^{\pm.00}$ & \cellcolor{blue!5}$0.001^{\pm.00}$ & \cellcolor{blue!5}$3.272^{\pm.01}$ & \cellcolor{blue!5}$9.002^{\pm.07}$ & \cellcolor{blue!5}-- & \cellcolor{blue!5}$0.014^{\pm.00}$ & \cellcolor{blue!5}$3.832^{\pm.00}$ \\
\cdashline{2-11}
& T2M~\cite{guo2022generating} & $0.222^{\pm.00}$ & $0.354^{\pm.00}$ & $0.452^{\pm.00}$ & $\underline{3.416}^{\pm.07}$ & $5.436^{\pm.02}$ & $\underline{8.018}^{\pm.07}$ & $0.905^{\pm.03}$ & $1.857^{\pm.10}$ & $115.6^{\pm1.7}$ \\
& InterGen~\cite{liang2024intergen} & $0.393^{\pm.00}$ & $0.571^{\pm.00}$ & $0.671^{\pm.00}$ & $5.762^{\pm.30}$ & $4.564^{\pm.03}$ & $10.45^{\pm.09}$ & $\textbf{4.566}^{\pm.13}$ & $2.116^{\pm.13}$ & $77.09^{\pm3.9}$ \\
& InterMask~\cite{javed2024intermask} & $0.386^{\pm.00}$ & $0.565^{\pm.01}$ & $0.664^{\pm.00}$ & $6.720^{\pm.28}$ & $4.616^{\pm.03}$ & $10.61^{\pm.11}$ & $2.493^{\pm.12}$ & $2.549^{\pm.16}$ & $59.41^{\pm.69}$ \\
\cdashline{2-11}
& \cellcolor{blue!5}Interact2Ar* & $\underline{0.426}^{\pm.00}$ & $\underline{0.611}^{\pm.00}$ & $\underline{0.710}^{\pm.00}$ & $5.728^{\pm.29}$ & $\underline{4.190}^{\pm.02}$ & $10.67^{\pm.10}$ & $\underline{3.967}^{\pm.13}$ & $2.040^{\pm.17}$ & $53.30^{\pm.51}$ \\
& \cellcolor{blue!10}Interact2Ar & $\textbf{0.469}^{\pm.00}$ & $\textbf{0.672}^{\pm.00}$ & $\textbf{0.779}^{\pm.00}$ & $\textbf{0.352}^{\pm.01}$ & $\textbf{3.173}^{\pm.01}$ & $\textbf{9.271}^{\pm.08}$ & $1.421^{\pm.04}$ & $\textbf{0.123}^{\pm.00}$ & $\textbf{6.620}^{\pm.13}$ \\
\midrule
\multirow{6}{*}{\rotatebox{90}{Hands}} 
& \cellcolor{blue!5}\textit{Ground Truth} & \cellcolor{blue!5}$0.399^{\pm.00}$ & \cellcolor{blue!5}$0.597^{\pm.00}$ & \cellcolor{blue!5}$0.713^{\pm.00}$ & \cellcolor{blue!5}$0.002^{\pm.00}$ & \cellcolor{blue!5}$3.312^{\pm.01}$ & \cellcolor{blue!5}$8.370^{\pm.05}$ & \cellcolor{blue!5}-- & \cellcolor{blue!5}$0.017^{\pm.00}$ & \cellcolor{blue!5}$2.480^{\pm.00}$ \\
\cdashline{2-11}
& T2M~\cite{guo2022generating} & $0.196^{\pm.00}$ & $0.317^{\pm.00}$ & $0.405^{\pm.01}$ & $9.424^{\pm.12}$ & $5.398^{\pm.02}$ & $6.996^{\pm.11}$ & $0.791^{\pm.04}$ & $1.767^{\pm.09}$ & $113.1^{\pm1.9}$ \\
& InterGen~\cite{liang2024intergen} & $0.357^{\pm.00}$ & $0.536^{\pm.01}$ & $0.645^{\pm.00}$ & $1.389^{\pm.03}$ & $3.806^{\pm.02}$ & $8.006^{\pm.07}$ & $\textbf{3.736}^{\pm.11}$ & $2.049^{\pm.12}$ & $70.47^{\pm2.9}$ \\
& InterMask~\cite{javed2024intermask} & $0.360^{\pm.00}$ & $0.538^{\pm.00}$ & $0.647^{\pm.00}$ & $1.960^{\pm.05}$ & $3.794^{\pm.02}$ & $7.939^{\pm.07}$ & $1.895^{\pm.09}$ & $2.201^{\pm.16}$ & $59.77^{\pm.76}$ \\
\cdashline{2-11}
& \cellcolor{blue!5}Interact2Ar* & $\underline{0.382}^{\pm.00}$ & $\underline{0.566}^{\pm.00}$ & $\underline{0.678}^{\pm.00}$ & $\underline{0.906}^{\pm.02}$ & $\underline{3.563}^{\pm.01}$ & $\textbf{8.273}^{\pm.10}$ & $\underline{3.206}^{\pm.09}$ & $1.923^{\pm.09}$ & $51.03^{\pm.49}$ \\
& \cellcolor{blue!10}Interact2Ar & $\textbf{0.422}^{\pm.00}$ & $\textbf{0.629}^{\pm.00}$ & $\textbf{0.745}^{\pm.00}$ & $\textbf{0.257}^{\pm.01}$ & $\textbf{3.111}^{\pm.01}$ & $\underline{8.614}^{\pm.08}$ & $1.439^{\pm.05}$ & $\textbf{0.120}^{\pm.00}$ & $\textbf{7.474}^{\pm.16}$ \\
\bottomrule
\end{tabular}
\vspace{-0.2cm}
\caption{Comparison of our model (Interact2Ar) to the state of the art in human-human interaction motion generation on the Inter-X dataset. *Interact2Ar model is the version without autoregressive generation. All evaluations have been executed 20 times to elude the randomness of the generation. $\pm$ indicates the 95\% confidence interval. We highlight the \textbf{best} and the \underline{second best} results.}
\label{tab:results}
\vspace{-0.2cm}
\end{table*}

\subsection{Ablation Study}
\vspace{-2mm}
\cref{tab:ablation} shows an ablation study of different autoregressive configurations in terms of memory size. We observe that models without Mixed Memory degrade in performance as more memory is added, while those with Mixed Memory maintain better performance with larger contexts. The increase in context, while adding more information to the model, also increases the complexity that the model must learn. Based on benefits obtained in LLMs from larger contexts~\cite{liu2025comprehensive}, having a method that enables large context windows without significant memory overhead is promising. Based on the overall metrics, memory utilization, and qualitative examples, we selected the hyperparameters $m_s = 15$, $m_l = 45$, and $\delta = 5$, which provide a 60-frame context window while using only 24 frames. Additional metrics and experiments are available in the Supplementary Materials.

\begin{figure*}[!htpb]
  \centering
    \includegraphics[width=0.9\textwidth]{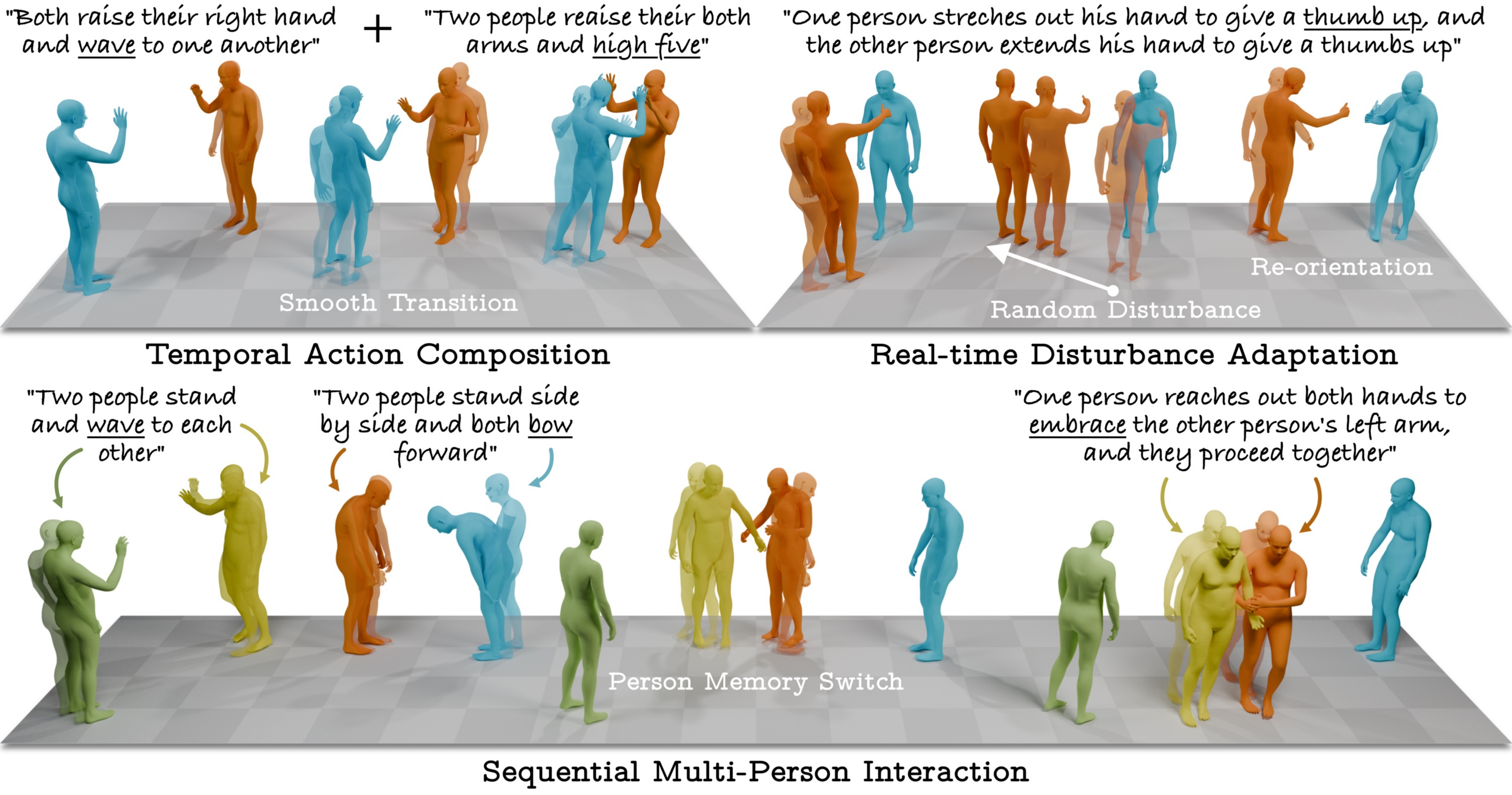}
    \vspace{-0.2cm}
    \caption{\textbf{Interact2Ar Adaptive Interactions.} Our method with autoregressive diffusion and a Mixed Memory strategy enables a series of adaptive downstream tasks, including temporal motion composition,  displacement adaptation, and sequential multi-human interactions.}
  \label{fig:qualitative:applications}
  \vspace{-0.3cm}
\end{figure*}

\begin{table}[!htbp]
\centering
\footnotesize
\setlength{\tabcolsep}{2 pt}
\begin{tabular}{c|c|c|ccccc}
\toprule
$m_s$ & $m_l$ & $\mathcal{M}$ & R-Prec. $\uparrow$ & FID $\downarrow$ & MM Dist. $\downarrow$ & Div. $\rightarrow$ & MMod. $\uparrow$ \\
\midrule
\cellcolor{blue!5}-&\cellcolor{blue!5}-& \cellcolor{blue!5}-- & \cellcolor{blue!5}$.722^{\pm.00}$ & \cellcolor{blue!5}$.001^{\pm.00}$ & \cellcolor{blue!5}$3.34^{\pm.00}$ & \cellcolor{blue!5}$9.08^{\pm.00}$ & \cellcolor{blue!5}--\\
\midrule
15 & - & 15 & $.765^{\pm.00}$ & $.283^{\pm.00}$ & $3.14^{\pm.01}$ & $9.28^{\pm.08}$ & $\textbf{1.49}^{\pm.05}$ \\
30 & - & 30 & $.769^{\pm.00}$ & $.346^{\pm.01}$ & $3.12^{\pm.01}$ & $\textbf{9.24}^{\pm.06}$ & $\underline{1.47}^{\pm.04}$ \\
60 & - & 60 & $\textbf{.776}^{\pm.00}$ & $.316^{\pm.01}$ & $\textbf{3.07}^{\pm.02}$ & $9.29^{\pm.07}$ & $1.39^{\pm.06}$ \\
90 & - & 90 & $.771^{\pm.00}$ & $.412^{\pm.01}$ & $3.10^{\pm.01}$ & $9.26^{\pm.07}$ & $1.34^{\pm.04}$ \\
120 & - & 120 & $\underline{.774}^{\pm.00}$ & $.413^{\pm.01}$ & $3.08^{\pm.01}$ & $9.30^{\pm.07}$ & $1.34^{\pm.04}$ \\
\cdashline{1-8}
15 & 15 & 18 & $.769^{\pm.00}$ & $.289^{\pm.01}$ & $3.13^{\pm.01}$ & $\underline{9.26}^{\pm.06}$ & $1.46^{\pm.05}$ \\
15 & 45 & 24 & $.773^{\pm.00}$ & $\textbf{.277}^{\pm.01}$ & $\underline{3.10}^{\pm.01}$ & $9.31^{\pm.07}$ & $1.43^{\pm.04}$ \\
15 & 75 & 30 & $.771^{\pm.00}$ & $\underline{.279}^{\pm.01}$ & $3.11^{\pm.01}$ & $9.32^{\pm.06}$ & $1.35^{\pm.05}$ \\
15 & 105 & 36 & $.773^{\pm.00}$ & $.325^{\pm.01}$ & $3.09^{\pm.01}$ & $9.35^{\pm.06}$ & $1.36^{\pm.04}$ \\
\bottomrule
\end{tabular}
\vspace{-0.2cm}
\caption{\textbf{Ablation study} on memory configurations for Interact2Ar. $m_s$ and $m_l$ represent the context window used for each memory. $m_l = -$ indicates models not using Mixed Memory. For models using Mixed Memory, $\delta = 5$. The total number of frames used in the full memory is $m = m_s + m_l / \delta$.}
\label{tab:ablation}
\vspace{-0.6cm}
\end{table}

\subsection{Adaptive Interactions}
\label{sec:experiments:additional}
As described in \cref{sec:method:applications}, Interact2Ar can seamlessly generate adaptive interactions thanks to our autoregressive diffusion model with Mixed Memory. We quantitatively evaluate this adaptability in terms of transition smoothness using the PJ and AUJ metrics as described at the beginning of this section. \cref{tab:results} shows that the autoregressive version of Interact2Ar clearly outperforms all previous methods. 

\cref{fig:qualitative:applications} shows our method seamlessly generating smooth temporal motion composition, adapting to random displacements, and handling sequential multi-human interactions. In the Supplementary video, these capabilities are more visually apparent, and comparisons with previous methods show that previous approaches produce highly abrupt transitions as a consequence of overfitting to initial positions.

%% file: sec/05_conclusion.tex
\section{Conclusion}
\label{sec:conclusion}
We introduced Interact2Ar, the first text-conditioned autoregressive diffusion model for generating full-body human-human interactions. Our cooperative denoiser architecture, with body-part specialized heads, achieves state-of-the-art performance on the Inter-X benchmark for full-body interaction generation, including detailed hand motions. We validated the quality of generated motions through quantitative and qualitative experiments, proposing a robust set of evaluators that better detect motion degradation and body-part-specific metrics that enable more informative assessment of interaction quality. Building on this foundation, we introduced an autoregressive diffusion pipeline that produces more aligned and adaptable interaction models, improving dyadic interaction generation while enabling downstream applications including temporal motion composition, disturbance adaptation, and multi-human interactions. Finally, our Mixed Memory paradigm leverages short- and long-term information to generate motions with broader temporal context while optimizing memory efficiency.

\noindent\textbf{Limitations and Future Work.} Interact2Ar introduces several contributions in full-body human-human interaction generation. However, some limitations originating from dataset constraints shape future work. Accounting for human diversity in realistic interactions requires modeling distinct body shapes. While Inter-X has largely improved over previous datasets in terms of full-body motions, shape distributions remain a challenge. Consequently, the dataset provides body shapes normalized to neutral, which hinders the precision of hand contacts between individuals.

\paragraph{Acknowledgments.} 
This work was partially supported by the Spanish national grant for PhD studies (FPU22/04200), the Spanish project PID2022-136436NB-I00, the ICREA Academia programme, the Valencian regional government CIPROM/2026/106 Prometeo group project AI-XCARE, and by the Spanish State Research Agency (AEI) and ERDF/EU under grant: GEMELIA PID2024-161711OB-I00. Additionally, we thank Julian Tanke for fruitful discussions regarding the limitations of current evaluators of human motion models

%% file: sec/99_suppl.tex
\clearpage
\maketitlesupplementary

This supplementary material aims to enhance the reproducibility and understanding of the work contributions. In \cref{supp:implementation}, we outline the implementation details of the state-of-the-art models used for comparison, provide detailed formulations of the loss functions employed in training, describe the newly proposed body-part-specific evaluators, detail the user study methodology, and explain the implementation of adaptive interaction capabilities. In \cref{supp:quantitative}, we complement the quantitative evaluation with results using the original evaluators from the Inter-X dataset, present an extended ablation study examining different memory configurations across all evaluation settings, evaluate the impact of different text encoders, and introduce additional dyadic-specific metrics to assess interaction quality and physical plausibility, including foot sliding. In \cref{supp:qualitative}, we describe the accompanying Supplementary Video, which includes additional visual examples and side-by-side comparisons with previous state-of-the-art methods, alongside new close-up visualizations of complex hand and body contacts, to better illustrate the Interact2Ar capabilities. Finally, in \cref{supp:code}, we detail the code and data availability to ensure full reproducibility of our work.

\section{Implementation Details}
\label{supp:implementation}

\subsection{State-of-the-art Implementations}
We compared Interact2Ar with previous SOTA methods on the Inter-X dataset. We primarily compared against T2M~\cite{guo2022generating}, InterGen~\cite{liang2024intergen}, and InterMask~\cite{javed2024intermask}.

InterGen is the state-of-the-art baseline among the original baselines proposed by the Inter-X~\cite{xu2024inter} dataset authors. However, given that the weights are not public, we retrained InterGen using the original implementation details described in the paper, including a transformer encoder with 8 blocks and 8 heads (latent dimension 512, feed-forward dimension 1024). The model uses 1000 steps with DDIM-50 sampling and was trained for 5000 epochs using EMA and AdamW. Following Inter-X we used a learning rate $1 \times 10^{-4}$ with weight decay $2 \times 10^{-5}$ and a  batch size of 128. Given that the motion representation that we use is not the same as in the InterHuman dataset~\cite{liang2024intergen}, we used our loss adaptations using forward kinematics to train the model.

\subsection{Losses}
In this section, we provide detailed formulations and explanations for each component of our training loss function.

\noindent\textbf{Representation Loss.} The representation loss $\mathcal{L}_{\text{repr}}$ directly measures the $\ell_2$ distance between the predicted and ground truth SMPL-X parameters~\cite{SMPL-X:2019} in their raw representation space:
\begin{equation}
\mathcal{L}_{\text{repr}}(x,\hat{x}) = ||x - \hat{x}||_2^2,
\end{equation}
where $x \in \mathbb{R}^{T \times D}$ represents the ground truth SMPL-X parameters across $T$ frames with dimensionality $D$, and $\hat{x}$ denotes the predicted parameters. This loss operates directly on the body pose parameters, hand articulations, and global trajectory, providing a direct supervision signal in the learned representation space.

\noindent\textbf{Root Orientation Loss.} The root orientation loss $\mathcal{L}_{\text{orient}}$ specifically penalizes errors in the global root orientation of each individual:
\begin{equation}
\mathcal{L}_{\text{orient}}(r,\hat{r}) = ||r_a - \hat{r}_a||_2^2 + ||r_b - \hat{r}_b||_2^2,
\end{equation}
where $r_a, r_b \in \mathbb{R}^{T \times 3}$ represent the ground truth root orientations for individuals $a$ and $b$ respectively, and $\hat{r}_a, \hat{r}_b$ are the corresponding predictions. This loss ensures that the global facing direction and body orientation of each person are accurately captured, which is crucial for modeling proper spatial relationships in interactions.

\subsubsection{Kinematic Losses}
We compute the following geometric losses through forward kinematics (FK), which converts SMPL-X parameters to 3D joint positions: $p = \text{FK}(x)$.

\noindent\textbf{Joint Position Loss.} The global joint position loss $\mathcal{L}_{\text{pos}}$ penalizes discrepancies in the predicted 3D locations of body joints:
\begin{equation}
\mathcal{L}_{\text{pos}}(p, \hat{p}) = ||p_a - \hat{p}_a||_2^2 + ||p_b - \hat{p}_b||_2^2,
\end{equation}
where $p_a, p_b \in \mathbb{R}^{T \times N_j \times 3}$ are the ground truth global joint positions for both individuals with $N_j$ joints per person, and $\hat{p}_a, \hat{p}_b$ are the corresponding predicted positions. This loss enforces spatial accuracy in the generated motions.

\noindent\textbf{Joint Velocity Loss.} To promote temporal smoothness and physical plausibility, we apply a velocity loss on the joint positions:
\begin{equation}
\mathcal{L}_{\text{vel}}(v, \hat{v}) = ||v_a - \hat{v}_a||_2^2 + ||v_b - \hat{v}_b||_2^2,
\end{equation}
where $v_t = p_t - p_{t-1}$ represents the joint velocities computed as the difference between consecutive frames. This loss discourages unnatural jittering and encourages smooth, realistic motion trajectories.

\noindent\textbf{Foot Contact Loss.} The foot contact loss $\mathcal{L}_{\text{foot}}$ reduces artifacts such as foot skating and floating:
\begin{equation}
\mathcal{L}_{\text{foot}}(f, \hat{f}) = \sum_{i \in \{\text{feet}\}} ||v_i \odot f_i||_2^2 + ||\hat{v}_i \odot \hat{f}_i||_2^2,
\end{equation}
where $v_i$ denotes the velocity of foot joint $i$, $f_i \in \{0,1\}^T$ is a binary contact indicator (1 when the foot is in contact with the ground, 0 otherwise), and $\odot$ represents element-wise multiplication. This loss penalizes foot motion when contact is detected, enforcing physical constraints.

\noindent\textbf{Pairwise Joint Distance Map Loss.} To capture the fine-grained spatial relationships between the two individuals, we introduce the distance map loss $\mathcal{L}_{\text{dist}}$:
\begin{equation}
\mathcal{L}_{\text{dist}}(d, \hat{d}) = ||(D(p_a, p_b) - D(\hat{p}_a, \hat{p}_b)) \odot M||_2^2,
\end{equation}
where $D(p_a, p_b) \in \mathbb{R}^{T \times N_j \times N_j}$ computes the pairwise Euclidean distance between all joints of individual $a$ and all joints of individual $b$:
\begin{equation}
D(p_a, p_b)_{i,j} = ||p_a^{(i)} - p_b^{(j)}||_2,
\end{equation}
with $p_a^{(i)}$ and $p_b^{(j)}$ denoting the positions of the $i$-th joint of person $a$ and $j$-th joint of person $b$, respectively. The binary mask $M \in \{0,1\}^{T \times N_j \times N_j}$ activates the loss only for joint pairs in close proximity in the ground truth, focusing supervision on spatial relationships that are most critical for realistic interactions. This ensures that the spatial proximity patterns between the two individuals match the ground truth, which is essential for generating realistic interactive behaviors such as handshakes, hugs, and other contact-based interactions.

\noindent\textbf{Loss Weighting.} The weighting coefficients $\{\lambda_{\text{repr}}, \lambda_{\text{orient}}, \lambda_{\text{pos}}, \lambda_{\text{vel}}, \lambda_{\text{foot}}, \lambda_{\text{dist}}\}$ are determined through grid search to balance the contribution of each loss term. These weights are calibrated to normalize the magnitude differences across loss components, ensuring that each term contributes meaningfully to the optimization process. The specific values used in our experiments are: $\lambda_{\text{repr}} = 1.0$, $\lambda_{\text{orient}} = 0.1$, $\lambda_{\text{pos}} = 1.0$, $\lambda_{\text{vel}} = 1.0$, $\lambda_{\text{foot}} = 1.0$, and $\lambda_{\text{dist}} = 0.5$.

\subsection{Evaluators}
\cref{sec:experimentation} introduced an improved evaluation pipeline over the original evaluator provided on the Inter-X dataset. This new pipeline better assesses interaction quality and provides more granular information. To achieve this, we introduced a set of new body-part-specific evaluators retrained to have deeper knowledge of the global information of the interactants.

For all evaluators, we used the architecture proposed in~\cite{guo2022generating}, where a motion and a text feature extractor are trained via contrastive learning, and these encoded representations are used to calculate the remaining metrics. Using this architecture, we trained 3 evaluators for 300 epochs at a learning rate of $1 \times 10^{-4}$ to generate feature vectors of size 512. The full evaluator was trained using information from all SMPL-X joints, the body evaluator using only the base SMPL~\cite{SMPL:2015} joints, and the hand evaluator using the additional 30 joints used for hands.

We additionally made the evaluators more robust, as demonstrated in \cref{tab:robustness} Based on the findings of~\cite{meng2025rethinking}, we decided to train an evaluator using only joint positions. Since the positioning between different individuals has great importance for interactions, we represented joint positions using global coordinates. These coordinates are calculated using a forward kinematic function on SMPL-X rotations predicted by our model.

\subsection{User Study}
\begin{figure}[!htbp]
  \centering
    \includegraphics[width=\linewidth]{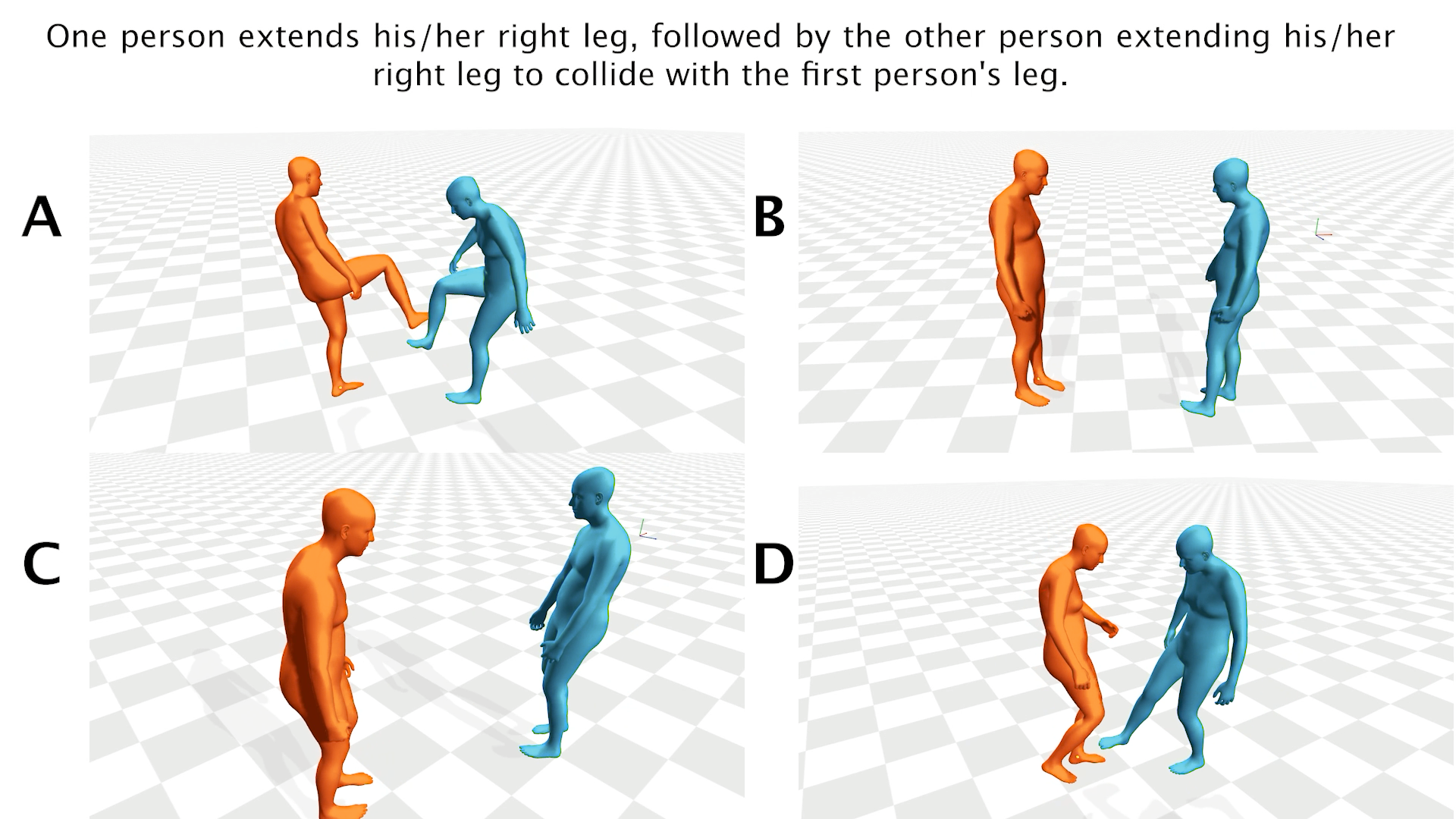}
    \caption{\textbf{User study.} A sample of the user study where 35 participants evaluated and ranked the text alignment and the hand quality of Interact2Ar and baseline methods.}  
    \label{fig:supp:userstudy}
\end{figure}

The user study was performed with 35 different participants to rank 10 different interactions extracted from: ground truth, our model, InterMask, and InterGen. The 35 participants were in the range of 25 to 55 years old, from different nationalities, all having higher degrees of study (bachelor's or more). Among the participants, there was a similar distribution of individuals familiarized with the human motion generation task and not. \cref{fig:supp:userstudy} presents a real frame from one of the videos that the users had to rank. In the video, there is a textual description at the top, and there are 4 videos randomly shuffled for each of the possible options. From each video, the participant had to rank each interaction based on the alignment with the textual description and the quality of the hand generation. We also included the ground truth provided by the dataset for this textual description, so the user always had an aligned interaction with the text and could rank all videos based on the overall quality. To ensure even distribution of the interaction motions, all the textual descriptions were extracted from the test set, and every one pertained to a different action category.

\begin{table*}[!htbp]
\centering
\small
\setlength{\tabcolsep}{3.5pt}
\begin{tabular}{c|c|ccccccc|cc}
\toprule
& \multirow{2}{*}{Methods} & \multicolumn{3}{c}{R-Precision $\uparrow$} & \multirow{2}{*}{FID $\downarrow$} & \multirow{2}{*}{MM Dist $\downarrow$} & \multirow{2}{*}{Diversity $\rightarrow$} & \multirow{2}{*}{MModality $\uparrow$} & \multirow{2}{*}{PJ $\uparrow$} & \multirow{2}{*}{AUJ $\downarrow$} \\
& & Top 1 & Top 2 & Top 3 & & & & & & \\
\midrule
\multirow{6}{*}{\rotatebox{90}{Full}} 
& \cellcolor{blue!5}\textit{Ground Truth} & \cellcolor{blue!5}$0.429^{\pm.00}$ & \cellcolor{blue!5}$0.626^{\pm.00}$ & \cellcolor{blue!5}$0.736^{\pm.00}$ & \cellcolor{blue!5}$0.002^{\pm.00}$ & \cellcolor{blue!5}$3.536^{\pm.01}$ & \cellcolor{blue!5}$9.734^{\pm.08}$ & \cellcolor{blue!5}-- & \cellcolor{blue!5}$0.021^{\pm.00}$ & \cellcolor{blue!5}$3.944^{\pm.00}$ \\
\cdashline{2-11}
& T2M~\cite{guo2022generating} & $0.184^{\pm.01}$ & $0.298^{\pm.01}$ & $0.396^{\pm.01}$ & $5.481^{\pm.38}$ & $9.576^{\pm.01}$ & $2.771^{\pm.15}$ & $2.761^{\pm.04}$ & $1.889^{\pm.12}$ & $124.9^{\pm1.9}$ \\
& InterGen~\cite{liang2024intergen} & $0.327^{\pm.00}$ & $0.506^{\pm.00}$ & $0.619^{\pm.00}$ & $2.601^{\pm.10}$ & $4.580^{\pm.02}$ & $\textbf{9.200}^{\pm.11}$ & $\textbf{3.999}^{\pm.11}$ & $2.132^{\pm.12}$ & $84.16^{\pm3.1}$ \\
& InterMask~\cite{javed2024intermask} & $0.403^{\pm.01}$ & $0.595^{\pm.00}$ & $0.705^{\pm.01}$ & $0.399^{\pm.01}$ & $3.705^{\pm.02}$ & $9.046^{\pm.07}$ & $2.261^{\pm.08}$ & $2.328^{\pm.21}$ & $61.74^{\pm.50}$ \\
\cdashline{2-11}
& \cellcolor{blue!5}Interact2Ar* & $\underline{0.433}^{\pm.00}$ & $\underline{0.623}^{\pm.00}$ & $\underline{0.727}^{\pm.00}$ & $\underline{0.255}^{\pm.01}$ & $\underline{3.618}^{\pm.01}$ & $9.066^{\pm.07}$ & $\underline{2.820}^{\pm.08}$ & $2.110^{\pm.21}$ & $54.97^{\pm.67}$ \\
& \cellcolor{blue!10}Interact2Ar & $\textbf{0.441}^{\pm.00}$ & $\textbf{0.631}^{\pm.00}$ & $\textbf{0.737}^{\pm.00}$ & $\textbf{0.148}^{\pm.01}$ & $\textbf{3.581}^{\pm.01}$ & $\underline{9.147}^{\pm.06}$ & $1.529^{\pm.05}$ & $\textbf{0.136}^{\pm.00}$ & $\textbf{8.837}^{\pm.17}$ \\
\midrule
\multirow{6}{*}{\rotatebox{90}{Body}} 
& \cellcolor{blue!5}\textit{Ground Truth} & \cellcolor{blue!5}$0.431^{\pm.00}$ & \cellcolor{blue!5}$0.621^{\pm.00}$ & \cellcolor{blue!5}$0.725^{\pm.00}$ & \cellcolor{blue!5}$0.002^{\pm.00}$ & \cellcolor{blue!5}$3.371^{\pm.01}$ & \cellcolor{blue!5}$8.931^{\pm.06}$ & \cellcolor{blue!5}-- & \cellcolor{blue!5}$0.014^{\pm.00}$ & \cellcolor{blue!5}$3.832^{\pm.00}$ \\
\cdashline{2-11}
& T2M~\cite{guo2022generating} & $0.310^{\pm.00}$ & $0.468^{\pm.00}$ & $0.570^{\pm.00}$ & $3.346^{\pm.05}$ & $4.465^{\pm.02}$ & $8.113^{\pm.09}$ & $0.604^{\pm.03}$ & $1.857^{\pm.10}$ & $115.6^{\pm1.7}$ \\
& InterGen~\cite{liang2024intergen} & $0.349^{\pm.00}$ & $0.528^{\pm.00}$ & $0.637^{\pm.00}$ & $1.708^{\pm.05}$ & $4.363^{\pm.02}$ & $9.253^{\pm.09}$ & $\textbf{3.596}^{\pm.11}$ & $2.116^{\pm.13}$ & $77.09^{\pm3.9}$ \\
& InterMask~\cite{javed2024intermask} & $0.401^{\pm.00}$ & $0.594^{\pm.00}$ & $0.707^{\pm.00}$ & $0.741^{\pm.03}$ & $3.483^{\pm.01}$ & $\textbf{8.915}^{\pm.08}$ & $1.589^{\pm.05}$ & $2.549^{\pm.16}$ & $59.41^{\pm.69}$ \\
\cdashline{2-11}
& \cellcolor{blue!5}Interact2Ar* & $\textbf{0.447}^{\pm.00}$ & $\textbf{0.643}^{\pm.00}$ & $\textbf{0.750}^{\pm.00}$ & $\underline{0.273}^{\pm.02}$ & $\textbf{3.231}^{\pm.01}$ & $9.074^{\pm.08}$ & $\underline{2.347}^{\pm.08}$ & $2.040^{\pm.17}$ & $53.30^{\pm.51}$ \\
& \cellcolor{blue!10}Interact2Ar & $\underline{0.446}^{\pm.00}$ & $\underline{0.639}^{\pm.00}$ & $\underline{0.744}^{\pm.00}$ & $\textbf{0.212}^{\pm.01}$ & $\underline{3.287}^{\pm.02}$ & $\underline{9.055}^{\pm.08}$ & $1.470^{\pm.05}$ & $\textbf{0.123}^{\pm.00}$ & $\textbf{6.620}^{\pm.13}$ \\
\midrule
\multirow{6}{*}{\rotatebox{90}{Hands}} 
& \cellcolor{blue!5}\textit{Ground Truth} & \cellcolor{blue!5}$0.372^{\pm.00}$ & \cellcolor{blue!5}$0.556^{\pm.00}$ & \cellcolor{blue!5}$0.663^{\pm.00}$ & \cellcolor{blue!5}$0.002^{\pm.00}$ & \cellcolor{blue!5}$3.893^{\pm.01}$ & \cellcolor{blue!5}$8.611^{\pm.07}$ & \cellcolor{blue!5}-- & \cellcolor{blue!5}$0.017^{\pm.00}$ & \cellcolor{blue!5}$2.480^{\pm.00}$ \\
\cdashline{2-11}
& T2M~\cite{guo2022generating} & $0.325^{\pm.00}$ & $0.486^{\pm.00}$ & $0.590^{\pm.00}$ & $2.114^{\pm.05}$ & $4.296^{\pm.02}$ & $8.129^{\pm.06}$ & $0.595^{\pm.04}$ & $1.767^{\pm.09}$ & $113.1^{\pm1.9}$ \\
& InterGen~\cite{liang2024intergen} & $0.331^{\pm.00}$ & $0.504^{\pm.00}$ & $0.617^{\pm.00}$ & $4.664^{\pm.19}$ & $4.560^{\pm.03}$ & $\textbf{9.654}^{\pm.12}$ & $\textbf{4.155}^{\pm.12}$ & $2.049^{\pm.12}$ & $70.47^{\pm2.9}$ \\
& InterMask~\cite{javed2024intermask} & $0.380^{\pm.00}$ & $0.568^{\pm.00}$ & $0.681^{\pm.00}$ & $0.383^{\pm.02}$ & $3.729^{\pm.02}$ & $\underline{8.689}^{\pm.08}$ & $1.883^{\pm.09}$ & $2.201^{\pm.16}$ & $59.77^{\pm.76}$ \\
\cdashline{2-11}
& \cellcolor{blue!5}Interact2Ar* & $\textbf{0.402}^{\pm.00}$ & $\textbf{0.592}^{\pm.00}$ & $\textbf{0.704}^{\pm.00}$ & $\underline{0.242}^{\pm.01}$ & $\textbf{3.639}^{\pm.02}$ & $8.972^{\pm.08}$ & $\underline{2.877}^{\pm.08}$ & $1.923^{\pm.09}$ & $51.03^{\pm.49}$ \\
& \cellcolor{blue!10}Interact2Ar & $\underline{0.393}^{\pm.00}$ & $\underline{0.584}^{\pm.00}$ & $\underline{0.695}^{\pm.00}$ & $\textbf{0.238}^{\pm.01}$ & $\underline{3.711}^{\pm.01}$ & $8.853^{\pm.07}$ & $1.754^{\pm.05}$ & $\textbf{0.120}^{\pm.00}$ & $\textbf{7.474}^{\pm.16}$ \\
\bottomrule
\end{tabular}
\vspace{-0.2cm}
\caption{Comparison of our model (Interact2Ar) to the state of the art in human-human interaction motion generation on the Inter-X dataset. *Interact2Ar model is the version without autoregressive generation. All evaluations have been executed 20 times to elude the randomness of the generation. $\pm$ indicates the 95\% confidence interval. We highlight the \textbf{best} and the \underline{second best} results.}
\label{tab:supp:results}
\vspace{-0.2cm}
\end{table*}

\subsection{Adaptive Interactions}
\noindent\textbf{Temporal Motion Composition.} We implement this with a large context window that includes all previously generated actions. Given the Mixed Memory approach we proposed, the memory buffer $\mathcal{M}$ accesses this information and enables Interact2Ar to generate seamless transitions while accounting for the complete action history.

\noindent\textbf{Real-Time Disturbance Adaptation.} To effectively assess the real-time adaptation of Interact2Ar, we randomly translated one individual in the XZ plane between different sub-motion generations. This simulates noisy contexts and disturbances produced by the environment or other individuals. Traditional diffusion models and Masked VQ-VAE Transformers, such as InterMask, cannot enable this capability because they produce the whole sequence at once, preventing adaptation until the entire motion is generated.

\noindent\textbf{Sequential Multi-Person Interactions.} We implemented this using two couples performing 2 different actions with their respective memories. Once they finish their actions, we take one individual from each couple and generate a new interaction with the newly formed couple. For the memories, we retain the original memories from the initial couples and create a new memory using information from the new couple. This enables seamless interaction while maintaining access to previously performed actions. While this implementation only generates sequential multi-human interactions, the idea can be expanded to generate parallel multi-human interactions as proposed in~\cite{xu2025multi}.

\section{Quantitative Evaluation}
\label{supp:quantitative}

\subsection{Original Evaluators} In \cref{tab:robustness}, we present a quantitative evaluation of the robustness of our newly proposed evaluators with respect to the original ones provided in the Inter-X dataset. Additionally, we provide the main metrics of our model in \cref{tab:results} using those newly trained evaluators. In \cref{tab:supp:results}, we performed the same evaluation using the original full-body evaluator trained directly with the SMPL-X representation, alongside body- and hand-specific evaluators using the same representation. We can observe in this case that Interact2Ar still outperforms previous methods. However, these differences are not as large as when using our evaluators. It can even be observed that in the body- and hand-specific evaluators, the non-autoregressive version of Interact2Ar obtains slightly better metrics than the autoregressive one. These smaller differences occur because rotation-based evaluators penalize diffusion models compared to VQ-VAE approaches. Furthermore, the original evaluators do not account for degradations in the global positioning of the individuals, which makes them incapable of detecting small differences, such as those between the autoregressive and non-autoregressive versions.

\begin{table*}[!htbp]
\centering
\small
\setlength{\tabcolsep}{3.5pt}
\begin{tabular}{c|c|ccc|ccccccc}
\toprule
& \multirow{2}{*}{Method} & \multirow{2}{*}{$m_s$} & \multirow{2}{*}{$m_l$} & \multirow{2}{*}{$\mathcal{M}$} & \multicolumn{3}{c}{R-Precision $\uparrow$} & \multirow{2}{*}{FID $\downarrow$} & \multirow{2}{*}{MM Dist $\downarrow$} & \multirow{2}{*}{Diversity $\rightarrow$} & \multirow{2}{*}{MModality $\uparrow$} \\
& & & & & Top 1 & Top 2 & Top 3 & & & & \\
\midrule
\multirow{11}{*}{\rotatebox{90}{Full}} 
& \cellcolor{blue!5}\textit{Ground Truth} & \cellcolor{blue!5}- & \cellcolor{blue!5}- & \cellcolor{blue!5}-- & \cellcolor{blue!5}$0.431^{\pm.00}$ & \cellcolor{blue!5}$0.631^{\pm.00}$ & \cellcolor{blue!5}$0.740^{\pm.00}$ & \cellcolor{blue!5}$0.002^{\pm.00}$ & \cellcolor{blue!5}$3.318^{\pm.01}$ & \cellcolor{blue!5}$8.973^{\pm.10}$ & \cellcolor{blue!5}-- \\
\cdashline{2-12}
& \multirow{5}{*}{\parbox{2.5cm}{\centering \textbf{Interact2Ar} Regular Memory}} & 15 & - & 15 & $0.449^{\pm.00}$ & $0.653^{\pm.00}$ & $0.765^{\pm.00}$ & $0.283^{\pm.01}$ & $3.141^{\pm.01}$ & $9.275^{\pm.08}$ & $\textbf{1.491}^{\pm.05}$ \\
& & 30 & - & 30 & $0.445^{\pm.00}$ & $0.656^{\pm.00}$ & $0.769^{\pm.00}$ & $0.346^{\pm.01}$ & $3.122^{\pm.01}$ & $\textbf{9.240}^{\pm.06}$ & $\underline{1.471}^{\pm.04}$ \\
& & 60 & - & 60 & $\textbf{0.458}^{\pm.00}$ & $\textbf{0.665}^{\pm.00}$ & $\textbf{0.776}^{\pm.00}$ & $0.316^{\pm.01}$ & $\textbf{3.071}^{\pm.02}$ & $9.285^{\pm.07}$ & $1.394^{\pm.06}$ \\
& & 90 & - & 90 & $0.452^{\pm.00}$ & $0.660^{\pm.01}$ & $0.771^{\pm.00}$ & $0.412^{\pm.01}$ & $3.103^{\pm.01}$ & $9.264^{\pm.07}$ & $1.344^{\pm.04}$ \\
& & 120 & - & 120 & $\underline{0.456}^{\pm.00}$ & $\textbf{0.665}^{\pm.00}$ & $\underline{0.774}^{\pm.00}$ & $0.413^{\pm.01}$ & $\underline{3.084}^{\pm.01}$ & $9.299^{\pm.07}$ & $1.336^{\pm.04}$ \\
\cdashline{2-12}
& \multirow{4}{*}{\parbox{2.5cm}{\centering \textbf{Interact2Ar} Mixed Memory}} & 15 & 15 & 18 & $0.448^{\pm.00}$ & $0.655^{\pm.00}$ & $0.769^{\pm.00}$ & $0.289^{\pm.01}$ & $3.131^{\pm.01}$ & $\underline{9.255}^{\pm.06}$ & $1.461^{\pm.05}$ \\
& & 15 & 45 & 24 & $0.453^{\pm.00}$ & $0.661^{\pm.00}$ & $0.773^{\pm.00}$ & $\textbf{0.277}^{\pm.01}$ & $3.095^{\pm.01}$ & $9.305^{\pm.07}$ & $1.427^{\pm.04}$ \\
& & 15 & 75 & 30 & $0.453^{\pm.00}$ & $0.661^{\pm.00}$ & $0.771^{\pm.00}$ & $\underline{0.279}^{\pm.01}$ & $3.110^{\pm.01}$ & $9.318^{\pm.06}$ & $1.346^{\pm.05}$ \\
& & 15 & 105 & 36 & $\textbf{0.458}^{\pm.00}$ & $\underline{0.664}^{\pm.00}$ & $0.773^{\pm.00}$ & $0.325^{\pm.01}$ & $3.089^{\pm.01}$ & $9.346^{\pm.06}$ & $1.363^{\pm.04}$ \\
\midrule
\multirow{11}{*}{\rotatebox{90}{Body}} 
& \cellcolor{blue!5}\textit{Ground Truth} & \cellcolor{blue!5}- & \cellcolor{blue!5}- & \cellcolor{blue!5}-- & \cellcolor{blue!5}$0.462^{\pm.01}$ & \cellcolor{blue!5}$0.655^{\pm.00}$ & \cellcolor{blue!5}$0.758^{\pm.00}$ & \cellcolor{blue!5}$0.001^{\pm.00}$ & \cellcolor{blue!5}$3.272^{\pm.01}$ & \cellcolor{blue!5}$9.002^{\pm.07}$ & \cellcolor{blue!5}-- \\
\cdashline{2-12}
& \multirow{5}{*}{\parbox{2.5cm}{\centering \textbf{Interact2Ar} Regular Memory}} & 15 & - & 15 & $0.466^{\pm.00}$ & $0.667^{\pm.00}$ & $0.774^{\pm.00}$ & $0.399^{\pm.01}$ & $3.194^{\pm.02}$ & $9.319^{\pm.09}$ & $\textbf{1.483}^{\pm.05}$ \\
& & 30 & - & 30 & $0.471^{\pm.00}$ & $0.672^{\pm.00}$ & $0.780^{\pm.00}$ & $0.428^{\pm.01}$ & $3.168^{\pm.01}$ & $\textbf{9.195}^{\pm.06}$ & $1.439^{\pm.05}$ \\
& & 60 & - & 60 & $\underline{0.474}^{\pm.01}$ & $\underline{0.677}^{\pm.00}$ & $\underline{0.781}^{\pm.00}$ & $0.397^{\pm.01}$ & $\underline{3.158}^{\pm.01}$ & $\underline{9.244}^{\pm.08}$ & $1.388^{\pm.05}$ \\
& & 90 & - & 90 & $0.471^{\pm.00}$ & $\textbf{0.678}^{\pm.00}$ & $\underline{0.781}^{\pm.00}$ & $0.475^{\pm.01}$ & $3.188^{\pm.01}$ & $9.260^{\pm.07}$ & $1.337^{\pm.04}$ \\
& & 120 & - & 120 & $0.470^{\pm.00}$ & $0.676^{\pm.00}$ & $\underline{0.781}^{\pm.00}$ & $0.463^{\pm.01}$ & $3.168^{\pm.01}$ & $9.311^{\pm.07}$ & $1.304^{\pm.04}$ \\
\cdashline{2-12}
& \multirow{4}{*}{\parbox{2.5cm}{\centering \textbf{Interact2Ar} Mixed Memory}} & 15 & 15 & 18 & $0.471^{\pm.00}$ & $0.676^{\pm.00}$ & $0.780^{\pm.00}$ & $0.354^{\pm.01}$ & $3.169^{\pm.01}$ & $9.268^{\pm.07}$ & $\underline{1.442}^{\pm.04}$ \\
& & 15 & 45 & 24 & $0.469^{\pm.00}$ & $0.672^{\pm.00}$ & $0.779^{\pm.00}$ & $\underline{0.352}^{\pm.01}$ & $3.173^{\pm.01}$ & $9.271^{\pm.08}$ & $1.421^{\pm.04}$ \\
& & 15 & 75 & 30 & $\underline{0.474}^{\pm.00}$ & $0.673^{\pm.00}$ & $0.778^{\pm.00}$ & $\textbf{0.331}^{\pm.01}$ & $3.175^{\pm.01}$ & $9.276^{\pm.06}$ & $1.362^{\pm.05}$ \\
& & 15 & 105 & 36 & $\textbf{0.477}^{\pm.00}$ & $\textbf{0.678}^{\pm.00}$ & $\textbf{0.782}^{\pm.00}$ & $0.411^{\pm.01}$ & $\textbf{3.141}^{\pm.01}$ & $9.321^{\pm.06}$ & $1.343^{\pm.04}$ \\
\midrule
\multirow{11}{*}{\rotatebox{90}{Hands}} 
& \cellcolor{blue!5}\textit{Ground Truth} & \cellcolor{blue!5}- & \cellcolor{blue!5}- & \cellcolor{blue!5}-- & \cellcolor{blue!5}$0.399^{\pm.00}$ & \cellcolor{blue!5}$0.597^{\pm.00}$ & \cellcolor{blue!5}$0.713^{\pm.00}$ & \cellcolor{blue!5}$0.002^{\pm.00}$ & \cellcolor{blue!5}$3.312^{\pm.01}$ & \cellcolor{blue!5}$8.370^{\pm.05}$ & \cellcolor{blue!5}-- \\
\cdashline{2-12}
& \multirow{5}{*}{\parbox{2.5cm}{\centering \textbf{Interact2Ar} Regular Memory}} & 15 & - & 15 & $0.414^{\pm.00}$ & $0.620^{\pm.00}$ & $0.734^{\pm.00}$ & $\textbf{0.206}^{\pm.01}$ & $3.165^{\pm.01}$ & $8.580^{\pm.07}$ & $\textbf{1.540}^{\pm.06}$ \\
& & 30 & - & 30 & $0.420^{\pm.01}$ & $0.627^{\pm.00}$ & $0.744^{\pm.00}$ & $0.302^{\pm.01}$ & $3.129^{\pm.01}$ & $\textbf{8.515}^{\pm.07}$ & $1.495^{\pm.06}$ \\
& & 60 & - & 60 & $\underline{0.425}^{\pm.00}$ & $0.630^{\pm.00}$ & $\textbf{0.748}^{\pm.00}$ & $0.273^{\pm.01}$ & $\underline{3.112}^{\pm.01}$ & $8.595^{\pm.07}$ & $1.405^{\pm.06}$ \\
& & 90 & - & 90 & $0.423^{\pm.01}$ & $0.630^{\pm.01}$ & $\underline{0.747}^{\pm.00}$ & $0.350^{\pm.01}$ & $3.126^{\pm.01}$ & $8.607^{\pm.07}$ & $1.354^{\pm.05}$ \\
& & 120 & - & 120 & $0.424^{\pm.00}$ & $\underline{0.634}^{\pm.01}$ & $\textbf{0.748}^{\pm.00}$ & $0.378^{\pm.01}$ & $\textbf{3.111}^{\pm.01}$ & $8.589^{\pm.08}$ & $1.370^{\pm.05}$ \\
\cdashline{2-12}
& \multirow{4}{*}{\parbox{2.5cm}{\centering \textbf{Interact2Ar} Mixed Memory}} & 15 & 15 & 18 & $0.421^{\pm.00}$ & $0.628^{\pm.00}$ & $0.743^{\pm.00}$ & $0.258^{\pm.01}$ & $3.135^{\pm.01}$ & $\underline{8.551}^{\pm.08}$ & $\underline{1.504}^{\pm.06}$ \\
& & 15 & 45 & 24 & $0.422^{\pm.00}$ & $0.629^{\pm.00}$ & $0.745^{\pm.00}$ & $0.257^{\pm.01}$ & $\textbf{3.111}^{\pm.01}$ & $8.614^{\pm.08}$ & $1.439^{\pm.05}$ \\
& & 15 & 75 & 30 & $0.424^{\pm.00}$ & $0.632^{\pm.00}$ & $0.745^{\pm.00}$ & $\underline{0.245}^{\pm.01}$ & $3.124^{\pm.01}$ & $8.608^{\pm.06}$ & $1.372^{\pm.05}$ \\
& & 15 & 105 & 36 & $\textbf{0.430}^{\pm.00}$ & $\textbf{0.635}^{\pm.00}$ & $\textbf{0.748}^{\pm.00}$ & $0.269^{\pm.01}$ & $3.114^{\pm.01}$ & $8.576^{\pm.07}$ & $1.384^{\pm.05}$ \\
\bottomrule
\end{tabular}
\vspace{-0.2cm}
\caption{\textbf{Ablation study on memory configurations} for Interact2Ar across different evaluation settings. $m_s$ and $m_l$ represent the context window used for each memory. $m_l = -$ indicates models not using Mixed Memory. For models using Mixed Memory, $\delta = 5$. The total number of frames used in the full memory is $\mathcal{M} = m_s + m_l / \delta$.}
\label{tab:sup:ablation}
\vspace{-0.4cm}
\end{table*}

\subsection{Extended Memory Ablation}
In \cref{tab:ablation}, we present an extended ablation study where different memory configurations have been tested to determine which provides the best trade-off between quality in terms of metrics and memory size. \cref{tab:sup:ablation} is an extended version of this ablation where all the different evaluators have been used. As can be observed, the overall tendency that we observed for the full evaluator remains consistent in the body and hands evaluators. While adding more memory can result in more informative generations, it also increases the complexity of the task that the denoiser has to learn, resulting in a non-linear improvement of metrics as the memory size increases. However, what is more noticeable is the significant increase in FID, which is generally used to determine motion quality, when using Mixed Memory.

\subsection{Text Encoder Ablation}
CLIP~\cite{radford2021learning} is used as the text encoder for the textual descriptions of the interactions, which are injected into the model as conditions. This decision was made to ensure consistency with previous works~\cite{liang2024intergen,javed2024intermask,ruiz2024in2in}. To evaluate the impact of a newer and more advanced encoder, \cref{tab:supp:text_encoder_ablation} presents a comparison against Qwen3-VL-Embedding-2B~\cite{qwen3vlembedding}. As can be observed, the results show minimal differences, which can likely be attributed to the limited diversity of textual descriptions present in the dataset.

\begin{table*}[!htbp]
\centering
\small
\setlength{\tabcolsep}{3.5pt}
\begin{tabular}{c|ccccc}
\toprule
Methods & R-Precision (Top 3) $\uparrow$ & FID $\downarrow$ & MM Dist $\downarrow$ & Diversity $\rightarrow$ & MModality $\uparrow$ \\
\midrule
\cellcolor{blue!5}\textit{Ground Truth} & \cellcolor{blue!5}$0.740^{\pm.00}$ & \cellcolor{blue!5}$0.002^{\pm.00}$ & \cellcolor{blue!5}$3.318^{\pm.01}$ & \cellcolor{blue!5}$8.973^{\pm.10}$ & \cellcolor{blue!5}-- \\
\cdashline{1-6}
CLIP~\cite{radford2021learning}& \textbf{0.773}$^{\pm.00}$ & \textbf{0.277}$^{\pm.01}$ & \textbf{3.095}$^{\pm.01}$ & \underline{9.305}$^{\pm.07}$ & \underline{1.427}$^{\pm.04}$ \\
Qwen3-VL-Embedding-2B~\cite{qwen3vlembedding}& \underline{0.728}$^{\pm.00}$ & \underline{0.389}$^{\pm.02}$ & \underline{3.361}$^{\pm.03}$ & \textbf{9.230}$^{\pm.14}$ & \textbf{1.637}$^{\pm.04}$ \\
\bottomrule
\end{tabular}
\vspace{-0.2cm}
\caption{\textbf{Ablation study on the text encoder} used to encode textual conditions in Interact2Ar. We compare CLIP and Qwen3-VL-Embedding-2B using the original Inter-X evaluation metrics. The \textbf{best} and \underline{second best} results are highlighted.}
\label{tab:supp:text_encoder_ablation}
\end{table*}

\subsection{Additional Interaction Metrics}
In addition to the standard metrics present in the Inter-X benchmark, \cref{tab:supp:interaction_metrics} presents supplementary metrics from related tasks to provide further insights into interaction quality. Specifically, Contact Frequency~\cite{siyao2024duolando} measures the ratio of frames where interactants are in contact, $\text{FID}_{\text{CD}}$~\cite{siyao2024duolando} computes the FID using a feature vector derived from the pairwise distances of joints, and Interaction Volume Penetration~\cite{li2024interdance} calculates the average penetration volume per sequence between the individuals involved in the interaction. As can be observed, our proposed method consistently achieves the best performance across these additional metrics.

\begin{table*}[!htbp]
\centering
\small
\setlength{\tabcolsep}{3.5pt}
\begin{tabular}{c|cccc}
\toprule
Methods & Contact Frequency $\rightarrow$ & $\text{FID}_{\text{CD}}$ $\downarrow$ & Interaction Volume Penetration $\downarrow$ & PFC $\downarrow$ \\
\midrule
\cellcolor{blue!5}\textit{Ground Truth} & \cellcolor{blue!5}$33.827^{\pm2.67}$ \hphantom{{\color{gray}(-19.0\%)}} & \cellcolor{blue!5}-- & \cellcolor{blue!5}$0.086^{\pm.00}$ & \cellcolor{blue!5}$0.120^{\pm.00}$ \\
\cdashline{1-5}
T2M~\cite{guo2022generating} & $47.975^{\pm2.52}$ {\color{gray}(+14.1\%)} & \underline{4.037}$^{\pm.46}$ & $0.768^{\pm.06}$ & $2.454^{\pm.19}$ \\
InterGen~\cite{liang2024intergen} & $14.867^{\pm1.47}$ {\color{gray}(-19.0\%)} & $4.999^{\pm.41}$ & $0.441^{\pm.02}$ & $4.834^{\pm.25}$ \\
InterMask~\cite{javed2024intermask} & \underline{21.881}$^{\pm2.01}$ {\color{gray}(-11.9\%)} & $5.593^{\pm.33}$ & \underline{0.437}$^{\pm.03}$ & \underline{0.339}$^{\pm.03}$ \\
\cellcolor{blue!10}Interact2Ar & \textbf{38.703}$^{\pm2.27}$ {\color{gray}(+4.9\%)} & \textbf{2.406}$^{\pm.55}$ & \textbf{0.360}$^{\pm.08}$ & \textbf{0.268}$^{\pm.02}$ \\
\bottomrule
\end{tabular}
\vspace{-0.2cm}
\caption{\textbf{Evaluation with additional dyadic-specific metrics.} Contact Frequency calculates the ratio of frames where interactants are in contact, $\text{FID}_{\text{CD}}$ computes the FID using features derived from pairwise joint distances, Interaction Volume Penetration measures the average penetration volume per sequence, and PFC evaluates the physical plausibility of foot contacts. The \textbf{best} and \underline{second best} results are highlighted.}
\label{tab:supp:interaction_metrics}
\vspace{-0.2cm}
\end{table*}

\noindent\textbf{Foot Sliding.} Qualitative examples indicate that all evaluated methods suffer from foot sliding, a limitation that could be addressed with additional data or post-processing. Nevertheless, to quantitatively measure the physical plausibility of foot contacts, we include the physical foot contact (PFC) score~\cite{tseng2023edge} in \cref{tab:supp:interaction_metrics}. The results demonstrate that our approach yields the most realistic foot movements.

\section{Qualitative Evaluation}
\label{supp:qualitative}
\begin{figure}[H]
  \vspace{-0.2cm}
  \centering
    \includegraphics[width=\columnwidth]{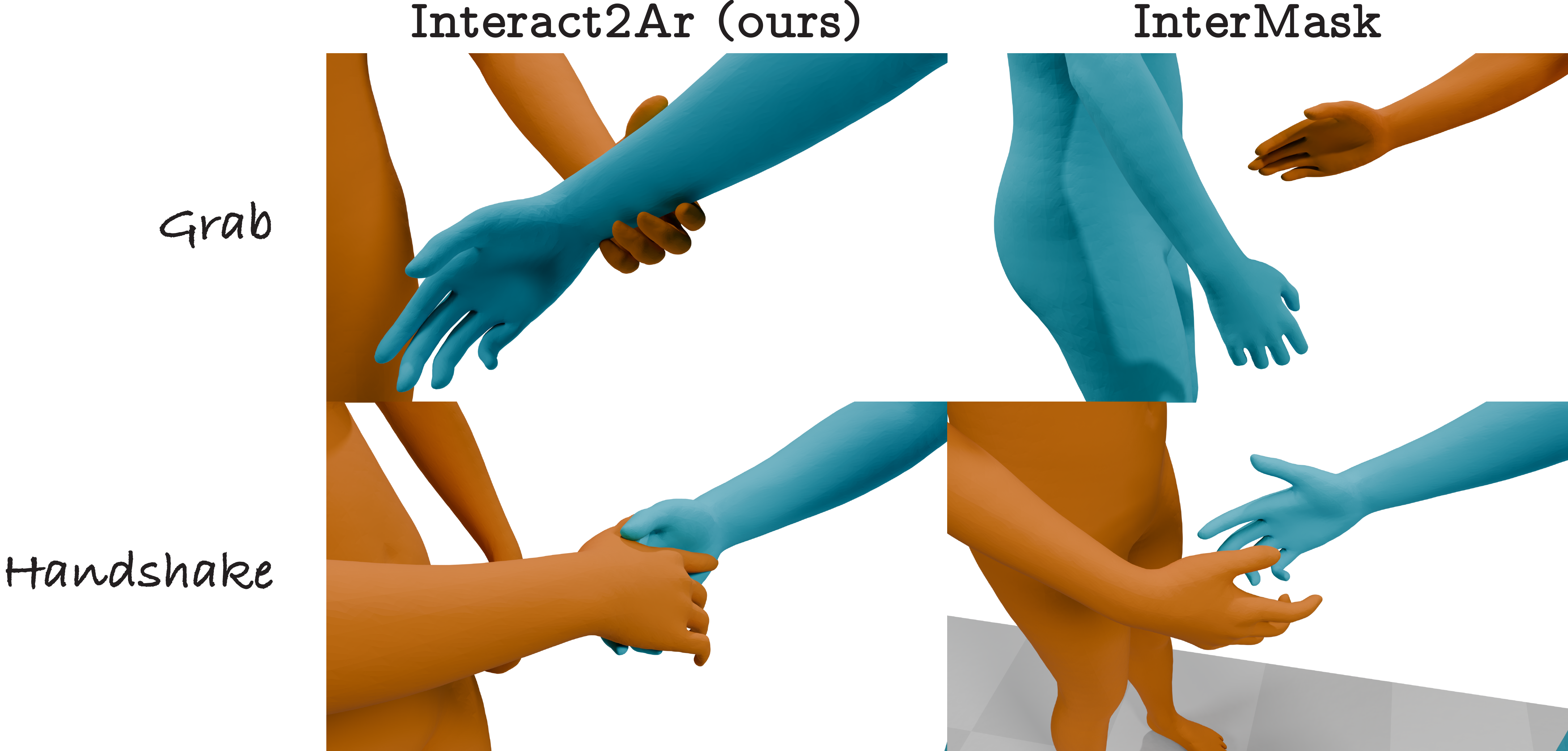}
  \caption{\textbf{Close-up visualizations.} Zoomed-in views of hands during challenging interactions involving body and hand contacts.}
  \label{fig:supp:close_hands}
  \vspace{-0.2cm}
\end{figure}

In addition to the qualitative examples shown in \cref{fig:qualitative:hhi} and \cref{fig:qualitative:applications}, we introduce a new set of examples and comparisons in the Supplementary Video. Due to the 4D nature of the representation that we generate, static images present a significant information loss. In the video, the quality of the motion and the side-by-side comparisons will facilitate understanding and highlight the qualitative differences between Interact2Ar and the previous SOTA, InterMask. However, to complement these dynamic results, \cref{fig:supp:close_hands} provides additional close-up static visualizations that further demonstrate the capability of Interact2Ar to generate realistic hand interactions involving complex body and hand contacts.

\section{Code and Data Availability}
\label{supp:code}
All the code and checkpoints related to this paper will be publicly released upon the acceptance of this paper. The code will contain all the codebase used to declare, train, and evaluate Interact2Ar and the new set of evaluators. The checkpoints will include the Interact2Ar checkpoints alongside the checkpoints of the evaluators for providing a more robust and reliable evaluation of future works using Inter-X.

%% file: main.bib
@String(CVPR= {IEEE Conf. Comput. Vis. Pattern Recog.})

@String(ICCV= {Int. Conf. Comput. Vis.})

@String(ECCV= {Eur. Conf. Comput. Vis.})

@String(TOG= {ACM Trans. Graph.})

@String(ICLR = {Int. Conf. Learn. Represent.})

@String(AAAI = {AAAI})

@String(CVPR  = {CVPR})

@String(ICCV  = {ICCV})

@String(ECCV  = {ECCV})

@String(TOG   = {ACM TOG})

@String(ICLR  = {ICLR})

@inproceedings{xu2024inter,
  title={Inter-x: Towards versatile human-human interaction analysis},
  author={Xu, Liang and Lv, Xintao and Yan, Yichao and Jin, Xin and Wu, Shuwen and Xu, Congsheng and Liu, Yifan and Zhou, Yizhou and Rao, Fengyun and Sheng, Xingdong and others},
  booktitle={Proceedings of the IEEE/CVF conference on computer vision and pattern recognition},
  pages={22260--22271},
  year={2024}
}

@article{liang2024intergen,
  title={Intergen: Diffusion-based multi-human motion generation under complex interactions},
  author={Liang, Han and Zhang, Wenqian and Li, Wenxuan and Yu, Jingyi and Xu, Lan},
  journal={International Journal of Computer Vision},
  volume={132},
  number={9},
  pages={3463--3483},
  year={2024},
  publisher={Springer}
}

@inproceedings{ruiz2024in2in,
  title={in2in: Leveraging individual information to generate human interactions},
  author={Ruiz-Ponce, Pablo and Barquero, German and Palmero, Cristina and Escalera, Sergio and Garc{\'\i}a-Rodr{\'\i}guez, Jos{\'e}},
  booktitle={Proceedings of the IEEE/CVF Conference on Computer Vision and Pattern Recognition},
  pages={1941--1951},
  year={2024}
}

@inproceedings{ruiz2025mixermdm,
  title={Mixermdm: Learnable composition of human motion diffusion models},
  author={Ruiz-Ponce, Pablo and Barquero, German and Palmero, Cristina and Escalera, Sergio and Garc{\'\i}a-Rodr{\'\i}guez, Jos{\'e}},
  booktitle={Proceedings of the Computer Vision and Pattern Recognition Conference},
  pages={12380--12390},
  year={2025}
}

@inproceedings{zhou2019continuity,
  title={On the continuity of rotation representations in neural networks},
  author={Zhou, Yi and Barnes, Connelly and Lu, Jingwan and Yang, Jimei and Li, Hao},
  booktitle={Proceedings of the IEEE/CVF conference on computer vision and pattern recognition},
  pages={5745--5753},
  year={2019}
}

@inproceedings{guo2022generating,
  title={Generating diverse and natural 3d human motions from text},
  author={Guo, Chuan and Zou, Shihao and Zuo, Xinxin and Wang, Sen and Ji, Wei and Li, Xingyu and Cheng, Li},
  booktitle={Proceedings of the IEEE/CVF conference on computer vision and pattern recognition},
  pages={5152--5161},
  year={2022}
}

@article{Plappert2016,
    author = {Matthias Plappert and Christian Mandery and Tamim Asfour},
    title = {The {KIT} Motion-Language Dataset},
    journal = {Big Data},
    publisher = {Mary Ann Liebert Inc},
    year = 2016,
    month = {dec},
    volume = {4},
    number = {4},
    pages = {236--252},
    url = {http://dx.doi.org/10.1089/big.2016.0028},
    doi = {10.1089/big.2016.0028},
}

@conference{AMASS:ICCV:2019,
  title = {{AMASS}: Archive of Motion Capture as Surface Shapes},
  author = {Mahmood, Naureen and Ghorbani, Nima and Troje, Nikolaus F. and Pons-Moll, Gerard and Black, Michael J.},
  booktitle = {International Conference on Computer Vision},
  pages = {5442--5451},
  month = oct,
  year = {2019},
  month_numeric = {10}
}

@article{h36m_pami,
author = {Ionescu, Catalin and Papava, Dragos and Olaru, Vlad and Sminchisescu,  Cristian},
title = {Human3.6M: Large Scale Datasets and Predictive Methods for 3D Human Sensing in Natural Environments},
journal = {IEEE Transactions on Pattern Analysis and Machine Intelligence},
publisher = {IEEE Computer Society},
volume = {36},
number = {7},
pages = {1325-1339},
month = {jul},
year = {2014}
}

@inproceedings{cai2022humman,
  title={{HuMMan}: Multi-modal 4d human dataset for versatile sensing and modeling},
  author={Cai, Zhongang and Ren, Daxuan and Zeng, Ailing and Lin, Zhengyu and Yu, Tao and Wang, Wenjia and Fan,
          Xiangyu and Gao, Yang and Yu, Yifan and Pan, Liang and Hong, Fangzhou and Zhang, Mingyuan and
          Loy, Chen Change and Yang, Lei and Liu, Ziwei},
  booktitle={17th European Conference on Computer Vision, Tel Aviv, Israel, October 23--27, 2022,
             Proceedings, Part VII},
  pages={557--577},
  year={2022},
  organization={Springer}
}

@article{xu2023towards,
  title={Towards continual egocentric activity recognition: A multi-modal egocentric activity dataset for continual learning},
  author={Xu, Linfeng and Wu, Qingbo and Pan, Lili and Meng, Fanman and Li, Hongliang and He, Chiyuan and Wang, Hanxin and Cheng, Shaoxu and Dai, Yu},
  journal={IEEE Transactions on Multimedia},
  volume={26},
  pages={2430--2443},
  year={2023},
  publisher={IEEE}
}

@inproceedings{shahroudy2016ntu,
  title={Ntu rgb+ d: A large scale dataset for 3d human activity analysis},
  author={Shahroudy, Amir and Liu, Jun and Ng, Tian-Tsong and Wang, Gang},
  booktitle={Proceedings of the IEEE conference on computer vision and pattern recognition},
  pages={1010--1019},
  year={2016}
}

@inproceedings{guo2020action2motion,
  title={Action2motion: Conditioned generation of 3d human motions},
  author={Guo, Chuan and Zuo, Xinxin and Wang, Sen and Zou, Shihao and Sun, Qingyao and Deng, Annan and Gong, Minglun and Cheng, Li},
  booktitle={Proceedings of the 28th ACM international conference on multimedia},
  pages={2021--2029},
  year={2020}
}

@inproceedings{BABEL:CVPR:2021,
  title = {{BABEL}: Bodies, Action and Behavior with English Labels},
  author = {Punnakkal, Abhinanda R. and Chandrasekaran, Arjun and Athanasiou, Nikos and Quiros-Ramirez, Alejandra and Black, Michael J.},
  booktitle = {Proceedings IEEE/CVF Conf.~on Computer Vision and Pattern Recognition (CVPR)},
  pages = {722--731},
  month = jun,
  year = {2021},
  doi = {},
  month_numeric = {6}
}

@inproceedings{yin2023hi4d,
      author = {Yin, Yifei and Guo, Chen and Kaufmann, Manuel and Zarate, Juan and Song, Jie and Hilliges, Otmar}, 
      title = {Hi4D: 4D Instance Segmentation of Close Human Interaction}, 
      booktitle = {Computer Vision and Pattern Recognition (CVPR)},
      year = {2023}
      }

@inproceedings{guo2021multi,
title={Multi-Person Extreme Motion Prediction},
author={Wen,Guo and Xiaoyu, Bie and Xavier, Alameda-Pineda and Francesc,Moreno-Noguer},
booktitle={Proceedings of the IEEE International Conference on Computer Vision and Pattern Recognition (CVPR)},
year={2022}
}

@inproceedings{fieraru2020three,
  title={Three-dimensional reconstruction of human interactions},
  author={Fieraru, Mihai and Zanfir, Mihai and Oneata, Elisabeta and Popa, Alin-Ionut and Olaru, Vlad and Sminchisescu, Cristian},
  booktitle={Proceedings of the IEEE/CVF Conference on Computer Vision and Pattern Recognition},
  pages={7214--7223},
  year={2020}
}

@article{ng2019you2me,
  title={You2Me: Inferring Body Pose in Egocentric Video via First and Second Person Interactions},
  author={Ng, Evonne and Xiang, Donglai and Joo, Hanbyul and Grauman, Kristen},
  journal={CVPR},
  year={2020}
}

@inproceedings{van2011umpm,
  title={Umpm benchmark: A multi-person dataset with synchronized video and motion capture data for evaluation of articulated human motion and interaction},
  author={Van der Aa, NP and Luo, Xinghan and Giezeman, Geert-Jan and Tan, Robby T and Veltkamp, Remco C},
  booktitle={2011 IEEE international conference on computer vision workshops (ICCV Workshops)},
  pages={1264--1269},
  year={2011},
  organization={IEEE}
}

@InProceedings{ghosh2024remos,
title={ReMoS: 3D Motion-Conditioned Reaction Synthesis for Two-Person Interactions},
author={Ghosh, Anindita and Dabral, Rishabh and Golyanik, Vladislav and Theobalt, Christian and Slusallek, Philipp},
booktitle={European Conference on Computer Vision (ECCV)},
year={2024}
}

@InProceedings{DSAG, author = {Gupta, Debtanu and Maheshwari, Shubh and Kalakonda, Sai Shashank and Manasvi and Sarvadevabhatla, Ravi Kiran}, title = {DSAG: A Scalable Deep Framework for Action-Conditioned Multi-Actor Full Body Motion Synthesis}, booktitle = {Proceedings of the IEEE/CVF Winter Conference on Applications of Computer Vision (WACV)}, month = {January}, year = {2023} }

@article{javed2024intermask,
  title={Intermask: 3d human interaction generation via collaborative masked modeling},
  author={Javed, Muhammad Gohar and Guo, Chuan and Cheng, Li and Li, Xingyu},
  journal={arXiv preprint arXiv:2410.10010},
  year={2024}
}

@inproceedings{tanke2023social,
  title={Social diffusion: Long-term multiple human motion anticipation},
  author={Tanke, Julian and Zhang, Linguang and Zhao, Amy and Tang, Chengcheng and Cai, Yujun and Wang, Lezi and Wu, Po-Chen and Gall, Juergen and Keskin, Cem},
  booktitle={Proceedings of the IEEE/CVF International Conference on Computer Vision},
  pages={9601--9611},
  year={2023}
}

@article{shafir2023human,
  title={Human motion diffusion as a generative prior},
  author={Shafir, Yonatan and Tevet, Guy and Kapon, Roy and Bermano, Amit H},
  journal={arXiv preprint arXiv:2303.01418},
  year={2023}
}

@inproceedings{cai2024digital,
  title={Digital life project: Autonomous 3d characters with social intelligence},
  author={Cai, Zhongang and Jiang, Jianping and Qing, Zhongfei and Guo, Xinying and Zhang, Mingyuan and Lin, Zhengyu and Mei, Haiyi and Wei, Chen and Wang, Ruisi and Yin, Wanqi and others},
  booktitle={Proceedings of the IEEE/CVF conference on computer vision and pattern recognition},
  pages={582--592},
  year={2024}
}

@article{yu2025socialgen,
  title={Socialgen: Modeling multi-human social interaction with language models},
  author={Yu, Heng and Zhang, Juze and Chen, Changan and Xiang, Tiange and Fang, Yusu and Niebles, Juan Carlos and Adeli, Ehsan},
  journal={arXiv preprint arXiv:2503.22906},
  year={2025}
}

@inproceedings{tanke2025dyadic,
  title={Dyadic Mamba: Long-term Dyadic Human Motion Synthesis},
  author={Tanke, Julian and Shibuya, Takashi and Uchida, Kengo and Saito, Koichi and Mitsufuji, Yuki},
  booktitle={Proceedings of the Computer Vision and Pattern Recognition Conference},
  pages={2868--2877},
  year={2025}
}

@inproceedings{xu2025multi,
  title={Multi-Person Interaction Generation from Two-Person Motion Priors},
  author={Xu, Wenning and Fan, Shiyu and Henderson, Paul and Ho, Edmond SL},
  booktitle={Proceedings of the Special Interest Group on Computer Graphics and Interactive Techniques Conference Conference Papers},
  pages={1--11},
  year={2025}
}

@article{wang2024intercontrol,
  title={InterControl: Zero-shot Human Interaction Generation by Controlling Every Joint},
  author={Wang, Zhenzhi and Wang, Jingbo and Li, Yixuan and Lin, Dahua and Dai, Bo},
  journal={Advances in Neural Information Processing Systems},
  volume={37},
  pages={105397--105424},
  year={2024}
}

@article{meng2025absolute,
  title={Absolute Coordinates Make Motion Generation Easy},
  author={Meng, Zichong and Han, Zeyu and Peng, Xiaogang and Xie, Yiming and Jiang, Huaizu},
  journal={arXiv preprint arXiv:2505.19377},
  year={2025}
}

@inproceedings{tevet2022motionclip,
  title={Motionclip: Exposing human motion generation to clip space},
  author={Tevet, Guy and Gordon, Brian and Hertz, Amir and Bermano, Amit H and Cohen-Or, Daniel},
  booktitle={European Conference on Computer Vision},
  pages={358--374},
  year={2022},
  organization={Springer}
}

@inproceedings{
tevet2023human,
title={Human Motion Diffusion Model},
author={Guy Tevet and Sigal Raab and Brian Gordon and Yoni Shafir and Daniel Cohen-or and Amit Haim Bermano},
booktitle={The Eleventh International Conference on Learning Representations },
year={2023},
url={https://openreview.net/forum?id=SJ1kSyO2jwu}
}

@inproceedings{kim2023flame,
  title={Flame: Free-form language-based motion synthesis \& editing},
  author={Kim, Jihoon and Kim, Jiseob and Choi, Sungjoon},
  booktitle={Proceedings of the AAAI Conference on Artificial Intelligence},
  volume={37},
  number={7},
  pages={8255--8263},
  year={2023}
}

@article{zhang2024motiondiffuse,
  title={Motiondiffuse: Text-driven human motion generation with diffusion model},
  author={Zhang, Mingyuan and Cai, Zhongang and Pan, Liang and Hong, Fangzhou and Guo, Xinying and Yang, Lei and Liu, Ziwei},
  journal={IEEE transactions on pattern analysis and machine intelligence},
  volume={46},
  number={6},
  pages={4115--4128},
  year={2024},
  publisher={IEEE}
}

@inproceedings{chen2023executing,
  title={Executing your commands via motion diffusion in latent space},
  author={Chen, Xin and Jiang, Biao and Liu, Wen and Huang, Zilong and Fu, Bin and Chen, Tao and Yu, Gang},
  booktitle={Proceedings of the IEEE/CVF conference on computer vision and pattern recognition},
  pages={18000--18010},
  year={2023}
}

@inproceedings{yuan2023physdiff,
  title={Physdiff: Physics-guided human motion diffusion model},
  author={Yuan, Ye and Song, Jiaming and Iqbal, Umar and Vahdat, Arash and Kautz, Jan},
  booktitle={Proceedings of the IEEE/CVF international conference on computer vision},
  pages={16010--16021},
  year={2023}
}

@inproceedings{zhang2023remodiffuse,
  title={Remodiffuse: Retrieval-augmented motion diffusion model},
  author={Zhang, Mingyuan and Guo, Xinying and Pan, Liang and Cai, Zhongang and Hong, Fangzhou and Li, Huirong and Yang, Lei and Liu, Ziwei},
  booktitle={Proceedings of the IEEE/CVF International Conference on Computer Vision},
  pages={364--373},
  year={2023}
}

@inproceedings{azadi2023make,
  title={Make-an-animation: Large-scale text-conditional 3d human motion generation},
  author={Azadi, Samaneh and Shah, Akbar and Hayes, Thomas and Parikh, Devi and Gupta, Sonal},
  booktitle={Proceedings of the IEEE/CVF International Conference on Computer Vision},
  pages={15039--15048},
  year={2023}
}

@inproceedings{zhou2024emdm,
  title={Emdm: Efficient motion diffusion model for fast and high-quality motion generation},
  author={Zhou, Wenyang and Dou, Zhiyang and Cao, Zeyu and Liao, Zhouyingcheng and Wang, Jingbo and Wang, Wenjia and Liu, Yuan and Komura, Taku and Wang, Wenping and Liu, Lingjie},
  booktitle={European Conference on Computer Vision},
  pages={18--38},
  year={2024},
  organization={Springer}
}

@inproceedings{huang2024stablemofusion,
  title={Stablemofusion: Towards robust and efficient diffusion-based motion generation framework},
  author={Huang, Yiheng and Yang, Hui and Luo, Chuanchen and Wang, Yuxi and Xu, Shibiao and Zhang, Zhaoxiang and Zhang, Man and Peng, Junran},
  booktitle={Proceedings of the 32nd ACM International Conference on Multimedia},
  pages={224--232},
  year={2024}
}

@inproceedings{guo2024momask,
  title={Momask: Generative masked modeling of 3d human motions},
  author={Guo, Chuan and Mu, Yuxuan and Javed, Muhammad Gohar and Wang, Sen and Cheng, Li},
  booktitle={Proceedings of the IEEE/CVF Conference on Computer Vision and Pattern Recognition},
  pages={1900--1910},
  year={2024}
}

@article{yuan2024mogents,
  title={Mogents: Motion generation based on spatial-temporal joint modeling},
  author={Yuan, Weihao and He, Yisheng and Shen, Weichao and Dong, Yuan and Gu, Xiaodong and Dong, Zilong and Bo, Liefeng and Huang, Qixing},
  journal={Advances in Neural Information Processing Systems},
  volume={37},
  pages={130739--130763},
  year={2024}
}

@inproceedings{pinyoanuntapong2024mmm,
  title={Mmm: Generative masked motion model},
  author={Pinyoanuntapong, Ekkasit and Wang, Pu and Lee, Minwoo and Chen, Chen},
  booktitle={Proceedings of the IEEE/CVF Conference on Computer Vision and Pattern Recognition},
  pages={1546--1555},
  year={2024}
}

@inproceedings{guo2022tm2t,
  title={Tm2t: Stochastic and tokenized modeling for the reciprocal generation of 3d human motions and texts},
  author={Guo, Chuan and Zuo, Xinxin and Wang, Sen and Cheng, Li},
  booktitle={European Conference on Computer Vision},
  pages={580--597},
  year={2022},
  organization={Springer}
}

@inproceedings{zhang2023generating,
  title={Generating human motion from textual descriptions with discrete representations},
  author={Zhang, Jianrong and Zhang, Yangsong and Cun, Xiaodong and Zhang, Yong and Zhao, Hongwei and Lu, Hongtao and Shen, Xi and Shan, Ying},
  booktitle={Proceedings of the IEEE/CVF conference on computer vision and pattern recognition},
  pages={14730--14740},
  year={2023}
}

@inproceedings{zhong2023attt2m,
  title={Attt2m: Text-driven human motion generation with multi-perspective attention mechanism},
  author={Zhong, Chongyang and Hu, Lei and Zhang, Zihao and Xia, Shihong},
  booktitle={Proceedings of the IEEE/CVF international conference on computer vision},
  pages={509--519},
  year={2023}
}

@article{jiang2023motiongpt,
  title={Motiongpt: Human motion as a foreign language},
  author={Jiang, Biao and Chen, Xin and Liu, Wen and Yu, Jingyi and Yu, Gang and Chen, Tao},
  journal={Advances in Neural Information Processing Systems},
  volume={36},
  pages={20067--20079},
  year={2023}
}

@inproceedings{zhang2024motiongpt,
  title={Motiongpt: Finetuned llms are general-purpose motion generators},
  author={Zhang, Yaqi and Huang, Di and Liu, Bin and Tang, Shixiang and Lu, Yan and Chen, Lu and Bai, Lei and Chu, Qi and Yu, Nenghai and Ouyang, Wanli},
  booktitle={Proceedings of the AAAI Conference on Artificial Intelligence},
  volume={38},
  number={7},
  pages={7368--7376},
  year={2024}
}

@inproceedings{
		tevet2025closd,
		title={{CL}o{SD}: Closing the Loop between Simulation and Diffusion for multi-task character control},
		author={Guy Tevet and Sigal Raab and Setareh Cohan and Daniele Reda and Zhengyi Luo and Xue Bin Peng and Amit Haim Bermano and Michiel van de Panne},
		booktitle={The Thirteenth International Conference on Learning Representations},
		year={2025},
		url={https://openreview.net/forum?id=pZISppZSTv}
}

@misc{zhang2025primalphysicallyreactiveinteractive,
      title={PRIMAL: Physically Reactive and Interactive Motor Model for Avatar Learning}, 
      author={Yan Zhang and Yao Feng and Alpár Cseke and Nitin Saini and Nathan Bajandas and Nicolas Heron and Michael J. Black},
      year={2025},
      eprint={2503.17544},
      archivePrefix={arXiv},
      primaryClass={cs.CV},
      url={https://arxiv.org/abs/2503.17544}, 
}

@inproceedings{camdm,
  title={Taming Diffusion Probabilistic Models for Character Control},
  author = {Chen, Rui and Shi, Mingyi and Huang, Shaoli and Tan, Ping and Komura, Taku and Chen, Xuelin},
  year = {2024},
  publisher = {Association for Computing Machinery},
  address = {New York, NY, USA},
  url = {https://doi.org/10.1145/3641519.3657440},
  doi = {10.1145/3641519.3657440},
  booktitle = {ACM SIGGRAPH 2024 Conference Papers},
  location = {Denver, CO, USA},
  series = {SIGGRAPH '24}
}

@article{shi2024interactive,
  title={Interactive character control with auto-regressive motion diffusion models},
  author={Shi, Yi and Wang, Jingbo and Jiang, Xuekun and Lin, Bingkun and Dai, Bo and Peng, Xue Bin},
  journal={ACM Transactions on Graphics (TOG)},
  volume={43},
  number={4},
  pages={1--14},
  year={2024},
  publisher={ACM New York, NY, USA}
}

@inproceedings{ji2025towards,
  title={Towards immersive human-x interaction: A real-time framework for physically plausible motion synthesis},
  author={Ji, Kaiyang and Shi, Ye and Jin, Zichen and Chen, Kangyi and Xu, Lan and Ma, Yuexin and Yu, Jingyi and Wang, Jingya},
  booktitle={Proceedings of the IEEE/CVF International Conference on Computer Vision},
  pages={10173--10183},
  year={2025}
}

@article{zhao2024dartcontrol,
  title={DartControl: A diffusion-based autoregressive motion model for real-time text-driven motion control},
  author={Zhao, Kaifeng and Li, Gen and Tang, Siyu},
  journal={arXiv preprint arXiv:2410.05260},
  year={2024}
}

@article{xiao2025motionstreamer,
  title={MotionStreamer: Streaming Motion Generation via Diffusion-based Autoregressive Model in Causal Latent Space},
  author={Xiao, Lixing and Lu, Shunlin and Pi, Huaijin and Fan, Ke and Pan, Liang and Zhou, Yueer and Feng, Ziyong and Zhou, Xiaowei and Peng, Sida and Wang, Jingbo},
  journal={arXiv preprint arXiv:2503.15451},
  year={2025}
}

@inproceedings{cen2025ready_to_react,
  title={Ready-to-React: Online Reaction Policy for Two-Character Interaction Generation},
  author={Cen, Zhi and Pi, Huaijin and Peng, Sida and Shuai, Qing and Shen, Yujun and Bao, Hujun and Zhou, Xiaowei and Hu, Ruizhen},
  booktitle={ICLR},
  year={2025}
}

@inproceedings{SMPL-X:2019,
  title = {Expressive Body Capture: {3D} Hands, Face, and Body from a Single Image},
  author = {Pavlakos, Georgios and Choutas, Vasileios and Ghorbani, Nima and Bolkart, Timo and Osman, Ahmed A. A. and Tzionas, Dimitrios and Black, Michael J.},
  booktitle = {Proceedings IEEE Conf. on Computer Vision and Pattern Recognition (CVPR)},
  pages     = {10975--10985},
  year = {2019}
}

@article{SMPL:2015,
      author = {Loper, Matthew and Mahmood, Naureen and Romero, Javier and Pons-Moll, Gerard and Black, Michael J.},
      title = {{SMPL}: A Skinned Multi-Person Linear Model},
      journal = {ACM Trans. Graphics (Proc. SIGGRAPH Asia)},
      month = oct,
      number = {6},
      pages = {248:1--248:16},
      publisher = {ACM},
      volume = {34},
      year = {2015}
    }

@inproceedings{xiao2022DDGAN,
    title={Tackling the Generative Learning Trilemma with Denoising Diffusion {GAN}s},
    author={Zhisheng Xiao and Karsten Kreis and Arash Vahdat},
    booktitle={International Conference on Learning Representations (ICLR)},
    year={2022}
}

@article{dhariwal2021diffusion,
  title={Diffusion models beat gans on image synthesis},
  author={Dhariwal, Prafulla and Nichol, Alexander},
  journal={Advances in neural information processing systems},
  volume={34},
  pages={8780--8794},
  year={2021}
}

@article{liu2025comprehensive,
  title={A comprehensive survey on long context language modeling},
  author={Liu, Jiaheng and Zhu, Dawei and Bai, Zhiqi and He, Yancheng and Liao, Huanxuan and Que, Haoran and Wang, Zekun and Zhang, Chenchen and Zhang, Ge and Zhang, Jiebin and others},
  journal={arXiv preprint arXiv:2503.17407},
  year={2025}
}

@article{ramesh2022hierarchical,
  title={Hierarchical text-conditional image generation with clip latents},
  author={Ramesh, Aditya and Dhariwal, Prafulla and Nichol, Alex and Chu, Casey and Chen, Mark},
  journal={arXiv preprint arXiv:2204.06125},
  volume={1},
  number={2},
  pages={3},
  year={2022}
}

@inproceedings{meng2025rethinking,
  title={Rethinking Diffusion for Text-Driven Human Motion Generation: Redundant Representations, Evaluation, and Masked Autoregression},
  author={Meng, Zichong and Xie, Yiming and Peng, Xiaogang and Han, Zeyu and Jiang, Huaizu},
  booktitle={Proceedings of the Computer Vision and Pattern Recognition Conference},
  pages={27859--27871},
  year={2025}
}

@inproceedings{barquero2024seamless,
  title={Seamless human motion composition with blended positional encodings},
  author={Barquero, German and Escalera, Sergio and Palmero, Cristina},
  booktitle={Proceedings of the IEEE/CVF Conference on Computer Vision and Pattern Recognition},
  pages={457--469},
  year={2024}
}

@article{zhu2023human,
  title={Human motion generation: A survey},
  author={Zhu, Wentao and Ma, Xiaoxuan and Ro, Dongwoo and Ci, Hai and Zhang, Jinlu and Shi, Jiaxin and Gao, Feng and Tian, Qi and Wang, Yizhou},
  journal={IEEE Transactions on Pattern Analysis and Machine Intelligence},
  volume={46},
  number={4},
  pages={2430--2449},
  year={2023},
  publisher={IEEE}
}

@article{sui2025survey,
  title={A survey on human interaction motion generation},
  author={Sui, Kewei and Ghosh, Anindita and Hwang, Inwoo and Zhou, Bing and Wang, Jian and Guo, Chuan},
  journal={arXiv preprint arXiv:2503.12763},
  year={2025}
}

@article{fan20253d,
  title={3D Human Interaction Generation: A Survey},
  author={Fan, Siyuan and Huang, Wenke and Cai, Xiantao and Du, Bo},
  journal={arXiv preprint arXiv:2503.13120},
  year={2025}
}

@article{qwen3vlembedding,
  title={Qwen3-VL-Embedding and Qwen3-VL-Reranker: A Unified Framework for State-of-the-Art Multimodal Retrieval and Ranking},
  author={Li, Mingxin and Zhang, Yanzhao and Long, Dingkun and Chen Keqin and Song, Sibo and Bai, Shuai and Yang, Zhibo and Xie, Pengjun and Yang, An and Liu, Dayiheng and Zhou, Jingren and Lin, Junyang},
  journal={arXiv preprint arXiv:2601.04720},
  year={2026}
}

@inproceedings{radford2021learning,
  title={Learning transferable visual models from natural language supervision},
  author={Radford, Alec and Kim, Jong Wook and Hallacy, Chris and Ramesh, Aditya and Goh, Gabriel and Agarwal, Sandhini and Sastry, Girish and Askell, Amanda and Mishkin, Pamela and Clark, Jack and others},
  booktitle={International conference on machine learning},
  pages={8748--8763},
  year={2021},
  organization={PmLR}
}

@inproceedings{tseng2023edge,
  title={Edge: Editable dance generation from music},
  author={Tseng, Jonathan and Castellon, Rodrigo and Liu, Karen},
  booktitle={Proceedings of the IEEE/CVF conference on computer vision and pattern recognition},
  pages={448--458},
  year={2023}
}

@article{siyao2024duolando,
  title={Duolando: Follower gpt with off-policy reinforcement learning for dance accompaniment},
  author={Siyao, Li and Gu, Tianpei and Yang, Zhitao and Lin, Zhengyu and Liu, Ziwei and Ding, Henghui and Yang, Lei and Loy, Chen Change},
  journal={arXiv preprint arXiv:2403.18811},
  year={2024}
}

@article{li2024interdance,
  title={Interdance: Reactive 3d dance generation with realistic duet interactions},
  author={Li, Ronghui and Zhang, Youliang and Zhang, Yachao and Zhang, Yuxiang and Su, Mingyang and Guo, Jie and Liu, Ziwei and Liu, Yebin and Li, Xiu},
  journal={arXiv preprint arXiv:2412.16982},
  year={2024}
}
